\documentclass[11pt]{article}

\AddToHook{package/hyperref/after}{%
  
}

\usepackage[final]{acl}

\usepackage[T1]{fontenc}

\usepackage[utf8]{inputenc}

\usepackage{microtype}

\usepackage{inconsolata}

\usepackage{graphicx}
\usepackage{pifont}
\DeclareRobustCommand{\benchmarkyes}{\textcolor{green!45!black}{\ding{51}}}
\DeclareRobustCommand{\benchmarkno}{\textcolor{red!70!black}{\ding{55}}}

\usepackage{placeins}

\title{Same Tasks, Different Apps: Why Mobile GUI Agents Fail to Generalize?}

\author{
  \textbf{Tien Tran}\textsuperscript{1*},
  \textbf{Namho Koh}\textsuperscript{1*},
  \textbf{Daiki E. Matsunaga}\textsuperscript{1},
  \textbf{Ayush Jain}\textsuperscript{2$\ddagger$},
  \textbf{Kee Eung Kim}\textsuperscript{1$\ddagger$\textdagger}
  \\
  \textsuperscript{1}KAIST
  \quad
  \textsuperscript{2}CMU
  \\
  \texttt{\{tien.tran,namhokoh,d.e.matsunaga,kekim\}@kaist.ac.kr}
  \quad
  \texttt{ayushj2@andrew.cmu.edu}
}

\begin{document}
\maketitle
\begingroup
\renewcommand{\thefootnote}{}
\footnotetext[1]{\textsuperscript{*}Equal contribution.}
\footnotetext{\textsuperscript{$\ddagger$}Equal advising.}
\footnotetext{\textsuperscript{\textdagger}Corresponding author.}
\endgroup
\begin{abstract} 
Mobile GUI agents deployed in real settings must work across different applications that support the same functionality. Most existing benchmarks test each task in only one app, so a high score can mean the agent understands the task, or only that it knows that particular app. We introduce \textbf{AnyAppBench}, a category-controlled live Android benchmark that evaluates cross-application generalization while keeping the user goal fixed. It spans 10 functional categories, 100 task templates, and 520 task--application pairs over 52 applications. Agents run from raw instructions and with app-independent sub-goals, and a VLM judge labels every failed run under a fixed failure taxonomy whose reliability is measured by human annotation. We find that, across 13 agents, success on the original application does not transfer reliably to new applications with the same goal. Furthermore, providing high-level sub-goal decomposition produces only small, category-dependent changes that do not close the gap, and the mix of failure types changes with the target interface. Based on those insights, we believe the AnyAppBench benchmark provides an important stepping stone toward robust real-world deployment of mobile GUI agents. Our code, data and the leaderboard can be found at the project website \url{https://anyappbench.github.io/}. 
\end{abstract}

\section{Introduction}

\label{sec:intro}

Mobile GUI agents have improved rapidly, with recent systems exceeding 90\%
task success on live Android benchmarks~\citep{kong2025mobileworld}. Yet
deployment requires more than success on one interface: an agent should send a
message, add a calendar event, or complete a to-do item across any application
that supports the same user goal.

Existing benchmarks do not isolate this capability. Offline datasets such as
Android-in-the-Wild~\citep{rawles2023aitw}, AndroidControl~\citep{li2024androidcontrol},
and AMEX~\citep{chai2025amex} evaluate fixed states or recorded trajectories.
Online benchmarks such as AndroidWorld~\citep{rawles2025androidworld},
AndroidLab~\citep{xu2025androidlab}, and SPA-Bench~\citep{chen2025spabench}
support live interaction but bind each task to one fixed app,
conflating task difficulty, agent strategy, and interface-specific adaptation.
Transfer-oriented benchmarks vary different axes: GUI-Odyssey~\citep{lu2024guiodyssey}
studies multi-application workflows, TransBench~\citep{lu2025transbench}
focuses on static GUI grounding, and FedMABench~\citep{wang2025fedmabench}
uses app heterogeneity for federated training. Thus, existing benchmarks do not
hold the functional task fixed while varying its application implementation.

\begin{figure*}[t]
    \centering
    \includegraphics[width=\textwidth]{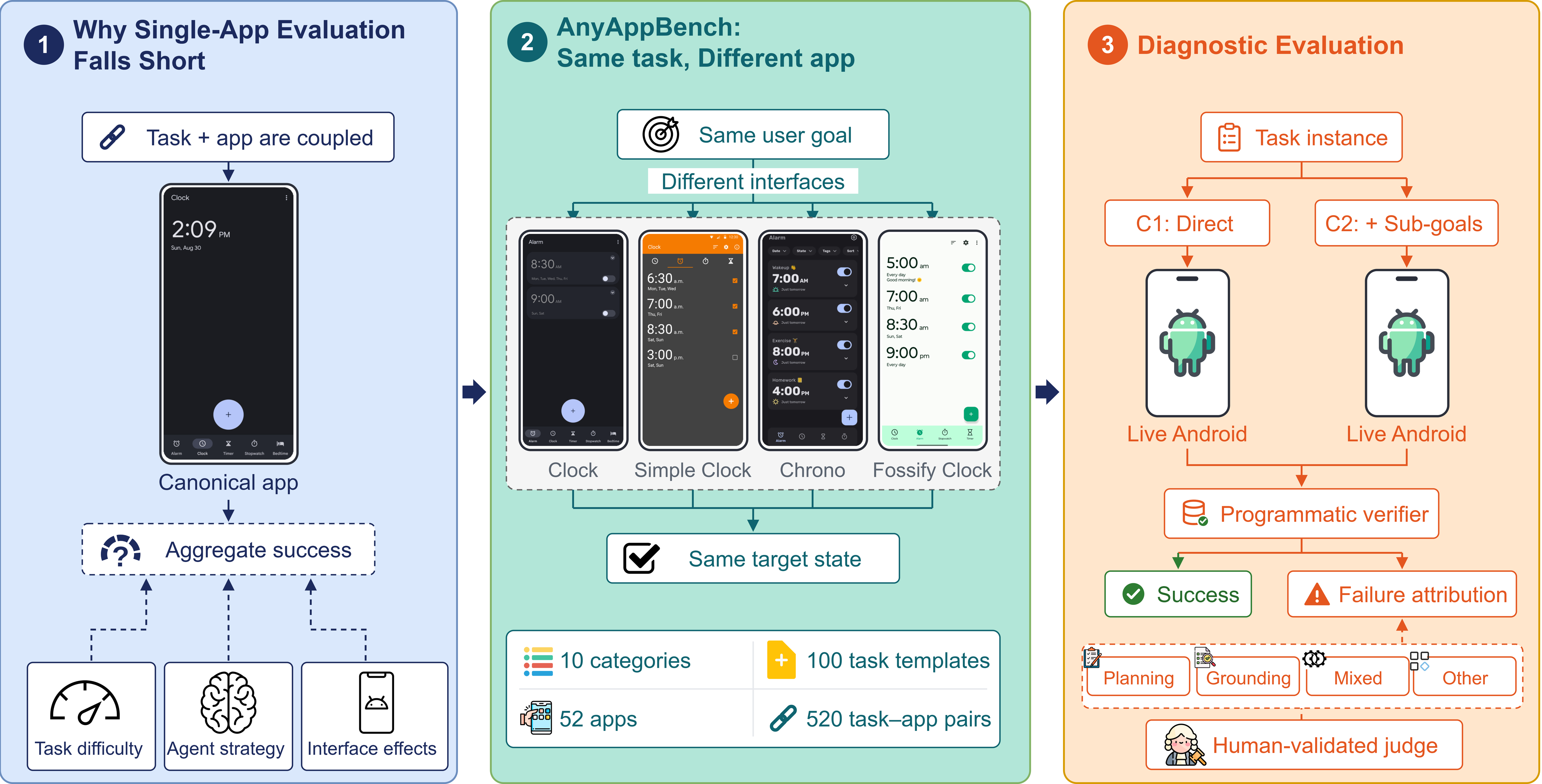}
    \caption{\textbf{Overview of AnyAppBench.} Single-application evaluation
    conflates task difficulty, agent strategy, and interface effects.
    AnyAppBench fixes the user goal and semantic success condition while varying
    the application within a functional category. Each task--application pair
    runs on live Android under direct (C1) and application-independent
    sub-goal-augmented (C2) conditions. Deterministic programmatic verifiers
    alone decide success. A VLM judge labels every verifier-failed C1
    trajectory under a fixed failure taxonomy, and a second judge plus human
    annotation measure how reliable those labels are. The benchmark spans 10 categories, 100 task templates, 52
    applications, and 520 validated task--application pairs.}
    \label{fig:AnyAppBench_overview}
\end{figure*}

To address this gap, we introduce \textbf{AnyAppBench}, a
category-controlled live Android benchmark for within-category
cross-application generalization. It defines reusable task templates with fixed
parameters, initialization logic, and semantic success conditions across 10 functional categories: Tasks, Notes, Finance, Music, Calendar, SMS, File
Manager, Maps, Contacts, and Clock. Each template is instantiated across
structurally diverse applications that support the same user-facing goal,
yielding 100 task templates, 52 runnable applications, and 520
task--application pairs. For diagnosis, AnyAppBench compares direct execution
with execution from externally generated, application-independent sub-goals.
A VLM judge attributes every failed run to a failure mode, and a second judge plus human annotation quantify how reliable those labels are.

Across 13 agents, AnyAppBench shows three key findings. First, the same task becomes much harder when only the app changes. With every application weighted equally, the 11 instruction-only agents succeed on 38.7\% of tasks on the original AndroidWorld apps but only on 27.3\% on new apps in the same category, a drop of 11.4 points. Therefore, a high score on the familiar app does not show that the agent has mastered the task. Second, the outcome depends strongly on which app is used, even though the goal is the same. The same goal can be easy in one app and difficult in another because layouts, labels, and workflows differ. Third, app-independent sub-goals do not close the gap. They raise success by 3.2 points on five categories and lower it by 3.5 points on the other five. Sub-goals tell the agent what to do, but they do not tell it where to find each control in an unfamiliar interface. Judge-based failure attribution further shows that the mix of failure types depends in part on the app, so diagnosis has to look at the interface as well as the task.

Our contributions are as follows:
\begin{itemize}
    \item We introduce \textbf{AnyAppBench}, a category-controlled benchmark that isolates within-category cross-application generalization by fixing the task and semantic terminal condition while varying the application.
    \item We build a live Android suite with 10 categories, 100 task templates, 52 applications, and 520 task--application pairs selected for interface diversity.
    \item We separate programmatic success verification from diagnosis: a VLM judge labels every failed run, and a second judge and human annotation measure the reliability of those labels.
    \item We evaluate 13 agents and show that mobile GUI performance and the
    observed composition of failures remain application-dependent even within
    a fixed functional category.
\end{itemize}
We make our code publicly available at \url{ https://anyappbench.github.io/}.
\section{Related Work}
\label{sec:related}
\paragraph{Mobile GUI agent benchmarks.}
Mobile GUI agents have motivated a growing set of datasets and benchmarks for screen understanding, action prediction, and closed-loop task execution. Offline resources such as Android-in-the-Wild~\citep{rawles2023aitw}, AndroidControl~\citep{li2024androidcontrol}, AMEX~\citep{chai2025amex}, Mobile-Bench~\citep{deng2024mobilebench}, Mobile-Bench-v2~\citep{xu2025mobilebenchv2}, MobileVLM/Mobile3M~\citep{wu2024mobilevlm}, MONDAY~\citep{jang2025monday}, ColorBench~\citep{song2025colorbench}, and MobiBench~\citep{im2025mobibench} provide screenshots, UI states, demonstrations, or multi-path trajectory annotations. Online benchmarks such as AndroidWorld~\citep{rawles2025androidworld}, AndroidLab~\citep{xu2025androidlab}, A3~\citep{chai2025a3}, SPA-Bench~\citep{chen2025spabench}, AndroidArena~\citep{xing2024androidarena}, and B-MoCA~\citep{lee2024bmoca} evaluate agents in live or emulated environments. AnyAppBench is complementary to these efforts: rather than adding more task types or longer workflows, it holds the functional task fixed and varies the application within the same category.

\paragraph{Cross-app and transfer-oriented evaluation.}
Several benchmarks study transfer or multi-app behavior, but they vary a different experimental axis from AnyAppBench. GUI-Odyssey~\citep{lu2024guiodyssey} collects trajectories that may involve multiple applications, making it useful for studying cross-app composition. TransBench~\citep{lu2025transbench} studies transfer across version, platform, and application axes, but its cross-application setting is primarily framed around static GUI grounding. FedMABench~\citep{wang2025fedmabench} constructs heterogeneous mobile-agent datasets for federated training and shows that app identity can be an important source of heterogeneity. AnyAppBench instead treats app variation as the controlled evaluation variable: the user instruction and intended final state are fixed, while the application interface changes.

\paragraph{Grounding and diagnostic evaluation.}
AnyAppBench also connects to work on GUI grounding and diagnostic agent evaluation. Grounding benchmarks and models such as SeeClick/ScreenSpot~\citep{cheng2024seeclick}, OS-Atlas~\citep{wu2024osatlas}, UGround~\citep{gou2025uground}, Set-of-Mark prompting~\citep{yang2023som}, and UI-Ins~\citep{chen2025uiins} study how models map language or visual targets to UI elements. Evaluation work such as LlamaTouch~\citep{zhang2024llamatouch}, GUIDE~\citep{zhai2026guide}, AgentRewardBench~\citep{lu2025agentrewardbench}, and Auto-Eval Judge~\citep{bhonsle2025autoeval} highlights the need to move beyond scalar success rates. AnyAppBench combines these threads: it evaluates the same category-level task across interchangeable apps, then uses human trajectory-level diagnosis to distinguish planning, grounding, mixed planning/grounding, execution/tooling, environment/evaluator, and unknown failures.

\section{AnyAppBench}
\label{sec:AnyAppBench}

AnyAppBench asks a simple question: if we keep the user's goal fixed and only change the app, does the agent still succeed? The benchmark has 100 task templates in 10 functional categories. Each template is paired with the app that AndroidWorld originally used for that category and with several new apps that offer the same function. This gives 520 validated task--application pairs across 11 original AndroidWorld apps and 41 new apps. Within a pair, the instruction, the parameters, and the condition for success stay the same. Only the app changes.

We build on top of AndroidWorld~\citep{rawles2025androidworld}. Forty-eight task intents and the 11 original apps come from it, and we reuse its emulator interface for screenshots, accessibility trees, access to persistent state, and low-level device actions. AnyAppBench adds 52 new task templates, 41 new apps, and the initialization and verification code needed to run each task on each app.

\subsection{Task Definition and Verification}
\label{sec:env}

An episode is defined by a task template $\tau$ and an app $\rho$. The template fixes the user instruction, its parameters, the initial device state, the condition that counts as success, and the step limit. The app determines the interface and the internal state through which the agent has to reach that condition.

\paragraph{Task equivalence.}
Two pairs are the same task when they share the goal, the parameters, the initial state, and the success condition. The apps may differ in layout, action sequence, wording, permissions, and storage. For example, ``set an alarm for 6:00 PM'' defines the same task across clock applications: success requires an enabled alarm at that time, regardless of the interface through which it is created or how the application stores alarm state. For each supported pair we write an app-specific verifier $v_{\tau,\rho}$ that reads the final state of app $\rho$ and checks the condition defined by $\tau$. The verifier translates one success criterion into each app's representation. It does not create a different task for each app.

\paragraph{Interaction space.}
At every step the agent sees a screenshot and a filtered accessibility tree taken from the same emulator state. It acts through six primitives: \texttt{tap(x,y)}, \texttt{long\_press(x,y)}, \texttt{swipe(x1,y1,x2,y2)}, \texttt{type(text)}, \texttt{key(keycode)}, and \texttt{wait(ms)}. An agent may plan at a higher level, but every command reaches the device through this interface~\citep{rawles2025androidworld}.

\paragraph{Programmatic verification.}
When the episode ends, $v_{\tau,\rho}$ returns a success or failure label that the agent never sees. As in AndroidWorld, a verifier may read persistent app state, the final screen or runtime state, or both~\citep{rawles2025androidworld}. We call these evidence sources durable-state (\textsc{D}), UI-state (\textsc{U}), and hybrid (\textsc{H}). The labels document where the evidence comes from; they do not change the decision rule. Appendix~\ref{app:verifier-taxonomy} lists verifier coverage.

\subsection{Benchmark Construction}
\label{sec:pipeline}
\begin{figure}[t]
\centering
\includegraphics[width=\columnwidth]{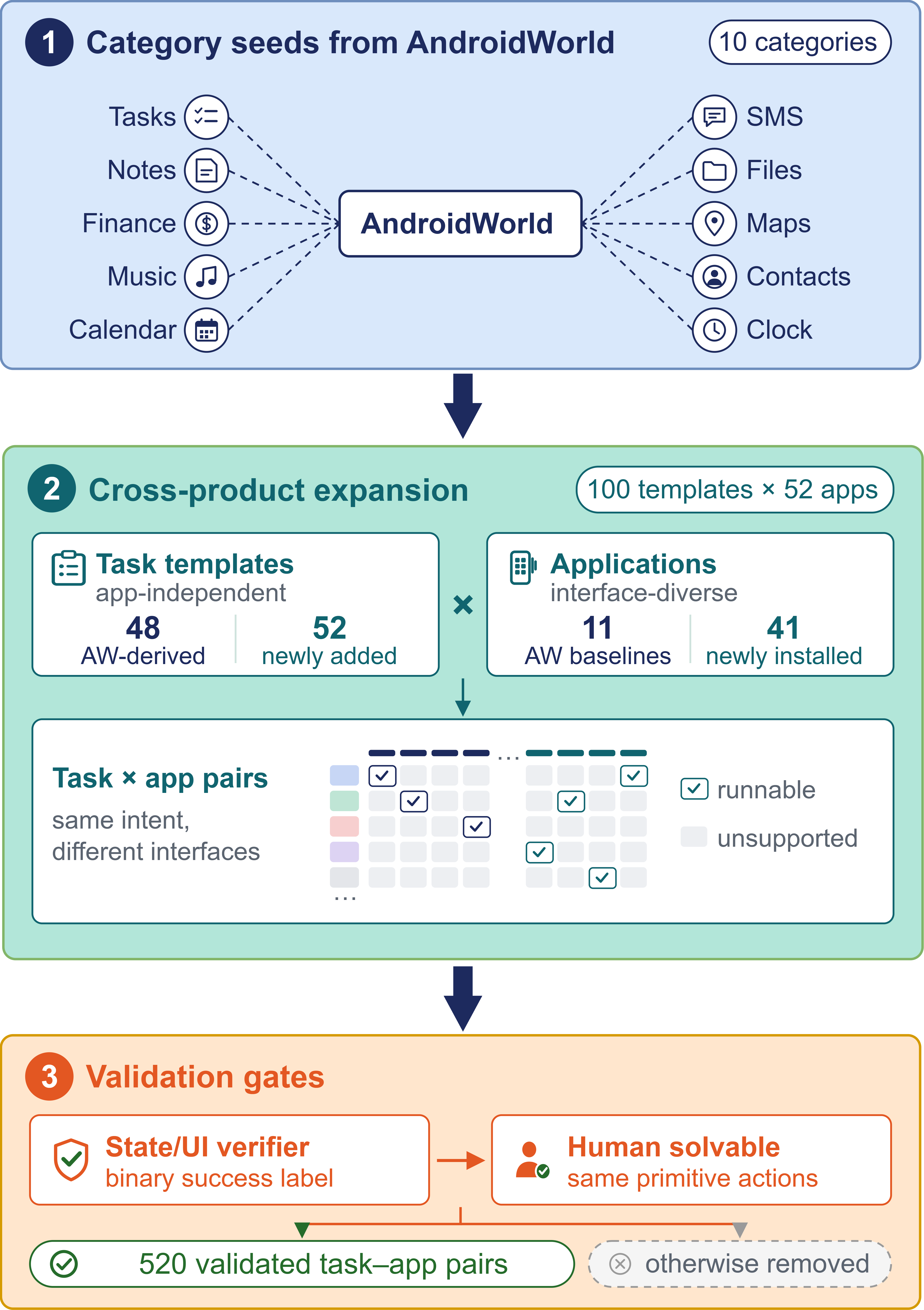}
\caption{\textbf{Construction of AnyAppBench.} We organize app-independent task templates into 10 functional categories and pair each category's 10 templates only with applications from the same category, producing 520 task--application pairs across 52 applications. A pair is retained only if it preserves the user-level goal, admits an equivalent programmatic verifier, runs end-to-end in the common harness, and is human-solvable using the agent's primitive action space.}
\label{fig:AnyAppBench_pipeline}
\end{figure}
Figure~\ref{fig:AnyAppBench_pipeline} shows the three construction stages.

\paragraph{Stage 1: Task templates.} The 10 categories are Tasks, Notes, Finance, Music, Calendar, SMS, File Manager, Maps, Contacts, and Clock. Each has 10 parameterized templates. Out of a total of 100 templates, 48 keep the intent of an AndroidWorld task~\citep{rawles2025androidworld}. The other 52 cover operations that AndroidWorld does not, such as making a task recurring (Tasks), forwarding a message thread (SMS), or adding a world clock (Clock). A template states its parameters, its initialization, and its success condition without referring to any particular app.

\paragraph{Stage 2: App expansion.} For each category we keep the original AndroidWorld app as the reference and add apps that support the same templates and can be operated through the six primitives. We profile every candidate on six interface properties (navigation depth, how standard its widgets are, home-screen density, naming convention, gesture convention, and number of primary screens). The profiles guide app selection only and play no role in initialization or verification. We drop near-duplicate interfaces because they add episodes without adding interface variation. All 11 original and 41 new apps were fixed before any agent was evaluated.

\paragraph{Stage 3: Pair validation.} A task--application pair is kept only if (i) the app supports the same user goal, (ii) an app-specific verifier can check the same success condition, (iii) initialization, execution, and verification run end to end in the shared harness, and (iv) a person can finish the task with the same six primitives the agent has. Pairs that change the requested outcome, lack the function, cannot be verified in the same way, or fail the human check are removed. The result is 520 pairs. The appendix lists all tasks, apps, interface profiles, and verifier metadata.

\subsection{Evaluation Tracks}
\label{sec:protocol}
Verifiers report whether an episode succeeded, not why it failed. Therefore, we complement task success with two diagnostic analyses: a controlled sub-goal intervention and trajectory-level failure attribution. Model coverage and run counts are stated in Section~\ref{sec:exp_setup}.

\paragraph{C1: Direct execution.} The agent receives the instruction and the current observation. It must infer the task structure, locate the relevant interface elements, execute the required actions, and decide when to terminate. C1 measures end-to-end success without external assistance.

\paragraph{C2: Sub-goal-augmented execution.} An external generator reads only the instruction and produces a short, application-independent list of sub-goals, which is prepended to the agent's prompt for the entire episode. The generator never observes the application, a screenshot, a trajectory, an action, or any feedback, and it provides no coordinates, accessibility identifiers, or application-specific labels. The agent must still locate and execute every step in the target interface. C1 and C2 share the same initialization and the same verifier. We interpret the comparison conservatively. A C1 failure that succeeds under C2 shows only that this particular decomposition changed the outcome, not that any valid decomposition would. A failure under both conditions is not informative on its own, because the supplied sub-goals may themselves be incomplete or incorrect.

\paragraph{Trajectory attribution.} Every C1 trajectory that fails verification receives one of six labels: planning, grounding, mixed planning/grounding, execution/tooling, environment/evaluator, or unknown. A VLM judge assigns the labels, an independent second judge re-labels the same trajectories, and stratified human annotation quantifies agreement with the judges. The two diagnostics are complementary: a planning label is a classification of a completed C1 trajectory into various failure modes, 
whereas the C2 comparison tests whether providing subgoal decomposition can explain the drop in performance.
Appendix~\ref{sec:human-protocol} describes the human protocol and Appendix~\ref{sec:judge-protocol} the automated judges and their agreement.

\section{Experiments and Results}
\label{sec:result}
We conduct a comprehensive evaluation of AnyAppBench to assess whether mobile GUI agents preserve task success when the application changes within the same functional category. AnyAppBench evaluates 13 baselines across 10 categories, 52 applications, and 520 task--application combinations. Each instance is evaluated under two execution conditions. C1 is direct execution. C2 prepends an application-independent sub-goal breakdown to the agent prompt.\\
Our analysis is driven by four research questions:

\begin{itemize}
    \item \textbf{RQ1.} How well does success on AndroidWorld-original applications transfer to newly installed same-category applications?
    \item \textbf{RQ2.} Is the cross-application gap driven only by AndroidWorld-inherited task templates?
    \item \textbf{RQ3.} Does app-independent sub-goal augmentation close the cross-application gap?
    \item \textbf{RQ4.} How does the observed composition of failure modes shift with the target application?
\end{itemize}
\subsection{Setup}
\label{sec:exp_setup}

\paragraph{Models.} We evaluate 13 baselines across five families (Table~\ref{tab:main_cross_app_gap_recomputed}): \textit{GUI-Trained} models post-trained for GUI control, \textit{General} vision--language models, \textit{Closed}-source proprietary models, \textit{Hybrid} configurations pairing a frontier planner with the UI-Venus-72B grounder, and \textit{Agent} systems wrapping a base model with planning, memory, and reflection.

\paragraph{Execution.} All evaluations use the same emulator,
six-primitive action space, and deterministic verifier. C1 covers all ten categories with three
runs per model--application--template cell. C2 runs on all ten categories with two runs per condition for the
five locally served agents, giving 5{,}165 matched C1/C2 rollout pairs.
Success is always determined by the programmatic verifier. Separately, a
two-stage Gemini judge assigns one of six diagnostic failure-mode labels to
all 4{,}529 verifier-failed trajectories of the audited five-category block, and these judge labels provide the
failure analysis in RQ4. A Qwen judge re-labels the same cases as a
judge-model sensitivity check, and 559 sampled trajectories receive human
labels that measure how reliable the judges are. Appendix~\ref{sec:human-protocol} describes the human
reference, and Appendix~\ref{sec:judge-protocol} reports the judge protocol
and agreement.

\paragraph{Metrics.} Let $\mathcal{R}_{\mathrm{can}}$ and
$\mathcal{R}_{\mathrm{alt}}$ denote the canonical and alternative applications,
respectively, and let $\mathrm{SR}_{\rho}$ be the success rate for application
$\rho$, expressed as a percentage. We first compute
\begin{align}
\overline{\mathrm{SR}}_{\mathrm{can}}
&= \frac{1}{|\mathcal{R}_{\mathrm{can}}|}
\sum_{\rho\in\mathcal{R}_{\mathrm{can}}}\mathrm{SR}_{\rho}, \notag\\
\overline{\mathrm{SR}}_{\mathrm{alt}}
&= \frac{1}{|\mathcal{R}_{\mathrm{alt}}|}
\sum_{\rho\in\mathcal{R}_{\mathrm{alt}}}\mathrm{SR}_{\rho}. \notag
\end{align}
The cross-application metrics are
\begin{align}
\Delta_{\mathrm{app}}
&= \overline{\mathrm{SR}}_{\mathrm{alt}}
 - \overline{\mathrm{SR}}_{\mathrm{can}}, \notag\\
\sigma_{\mathrm{app}}
&= \operatorname{std}_{\rho\in\mathcal{R}_{\mathrm{alt}}}
   (\mathrm{SR}_{\rho}). \notag
\end{align}
$\Delta_{\mathrm{app}}$ is the signed difference between the equal-app-weighted
success rates on alternative and canonical applications, reported in percentage
points; negative values indicate lower success on alternative applications.
$\sigma_{\mathrm{app}}$ is the population standard deviation of success rates
across alternative applications and measures application-to-application
variability.

We first average task templates and runs within each application and then weight
applications equally. When a category contains multiple canonical applications,
their success rates are averaged before computing $\Delta_{\mathrm{app}}$.
For the ten-category headline, $\overline{\mathrm{SR}}_{\mathrm{can}}$ and
$\overline{\mathrm{SR}}_{\mathrm{alt}}$ are computed as globally
equal-app-weighted means.

\paragraph{Matched inference.} For the matched C1/C2 analysis, the unit of pairing is the rollout cell: one (agent, application, task template, run-seed) combination evaluated once under C1 and once under C2 with the same seed and the same programmatic verifier (run~1 pairs C1's first seeded run with C2's first replicate; run~2 pairs the second of each). Cells that are infrastructure-invalid on either side are excluded, leaving 2{,}285 pairs. 95\% confidence intervals use a paired cluster bootstrap on the per-pair success differences: clusters are the 115 agent$\times$application groups (matching the clustering used for the judge agreement in Appendix~\ref{sec:judge-models}), resampled with replacement $B{=}10{,}000$ times (NumPy \texttt{default\_rng}, seed 20260831), with percentile intervals. The aggregate change of $+3.2$ points has CI $[+0.3, +6.1]$ ($+6.2$ on AndroidWorld-original apps, CI $[-2.0, +14.8]$; $+2.3$ on new apps, CI $[-0.6, +5.2]$). Reported $p$-values are two-sided exact McNemar tests over the discordant pairs (aggregate: 265 gained vs.\ 193 lost, $p<0.001$); per-agent, only UI-Venus-7B's gain excludes zero ($+9.6$, CI $[+3.7, +16.3]$).

\subsection{RQ1: Cross-Application Transfer}
\label{sec:rq1_cross_app}

Across all ten categories, the 11 instruction-only agents succeed on 38.7\%
of tasks on the original AndroidWorld apps and on 27.3\% on new apps, with
every app weighted equally, a signed change of $-11.4$ percentage points
(cell-level bootstrap CI $[-14.1,-9.1]$). Including the two hybrid systems,
the 13-agent average falls from 43.6\% to 31.4\%.

All 13 rows in Table~\ref{tab:main_cross_app_gap_recomputed} are negative for both template groups. Within the SMS, File Manager, Maps, Contacts, and Clock block, where three seeded runs support significance testing, the drop is significant ($p<0.05$, two-proportion $z$-test over rollouts) for 9 of the 11 instruction-only agents, and the original app beats the new apps on 39 of that block's 50 templates (10 reversed, 1 tie, sign test $p<0.001$). Models that do well on the original apps lose the most: on AW-inherited templates Gemini-3.1-Pro loses \textbf{29.0} points and UI Voyager-4B \textbf{25.3}, and GPT-5.1, MAI-UI-8B, and Qwen2.5-VL-72B each lose between \textbf{13} and \textbf{19}. The new AnyAppBench templates show the same pattern. Therefore, a high AndroidWorld score does not mean the agent has learned the task. New apps change layouts, labels, control positions, and workflows, and models that succeed on the original app often fail when only the interface changes. The large per-app standard deviations (typically 20--37 points) show that this is uneven across apps, not a uniform shift. Within the five-category block the gap is also stable across the three seeded runs (15.6, 15.3, and 18.6 points, per-run tables in the appendix). In this ten-category view the hybrid ($^\dagger$) systems no longer stand out: handing coordinate selection to a separate grounder helped on the five-category block, but over all ten categories their drops (\textbf{12.8}--\textbf{27.3} points) match the other agents. Appendix~\ref{app:full_results} gives per-category and per-app results. Figure~\ref{fig:qualitative_examples} in the appendix illustrates this failure mode: GPT-5.1 completes the same contact-management task in the original application but does not complete the save and favorite workflow in a new application.

\begin{table*}[t]
\centering
\footnotesize
\setlength{\tabcolsep}{3.5pt}
\renewcommand{\arraystretch}{1.05}
\caption{\textbf{Signed cross-app generalization change over all ten categories, decomposed by template provenance.}
\textbf{(a)} AW-inherited templates. \textbf{(b)} New AnyAppBench templates. \textbf{AW app}: equal-app success on the
original AndroidWorld apps. \textbf{New}: equal-app success on the new same-category apps. \textbf{Std}: population
std across new apps. $\boldsymbol{\Delta}$: New $-$ AW app.}
\label{tab:main_cross_app_gap_recomputed}
\begin{adjustbox}{max width=\textwidth}
\begin{tabular}{ll >{\columncolor{awcol}}c >{\columncolor{catcol}}c >{\columncolor{stdcol}}c >{\columncolor{deltacol}}c >{\columncolor{awcol}}c >{\columncolor{catcol}}c >{\columncolor{stdcol}}c >{\columncolor{deltacol}}c}
\toprule
& & \multicolumn{4}{c}{\textbf{(a) AW-inherited templates}} & \multicolumn{4}{c}{\textbf{(b) New AnyAppBench templates}} \\
\cmidrule(lr){3-6}\cmidrule(lr){7-10}
\textbf{Family} & \textbf{Model} & \textbf{AW app} & \textbf{New} & \textbf{Std} & $\boldsymbol{\Delta}$ & \textbf{AW app} & \textbf{New} & \textbf{Std} & $\boldsymbol{\Delta}$ \\
\midrule
\multirow{5}{*}{\rotatebox{90}{\scriptsize GUI Trained}} & UI-Venus-Navi-72B & 33.7 & 24.7 & 28.0 & -9.0 & 41.4 & 25.0 & 27.9 & -16.4 \\
 & UI-Venus-7B & 25.1 & 22.5 & 26.2 & -2.6 & 32.8 & 21.9 & 26.2 & -10.9 \\
 & GUI-Owl-7B & 31.6 & 19.5 & 27.2 & -12.1 & 21.6 & 16.7 & 20.1 & -4.9 \\
 & MAI-UI-8B & 42.5 & 28.8 & 31.9 & -13.7 & 39.8 & 28.7 & 30.1 & -11.1 \\
 & UI Voyager-4B & 53.9 & 28.6 & 26.1 & -25.3 & 23.2 & 19.1 & 23.2 & -4.1 \\
\midrule
\multirow{3}{*}{\rotatebox{90}{\scriptsize General}} & Qwen3-VL-30B-A3B & 21.7 & 18.5 & 26.8 & -3.2 & 35.2 & 16.5 & 24.6 & -18.7 \\
 & Qwen3-VL-8B & 20.7 & 17.7 & 26.5 & -3.0 & 21.4 & 15.7 & 21.8 & -5.7 \\
 & Qwen2.5-VL-72B & 38.4 & 25.3 & 27.7 & -13.0 & 40.0 & 27.4 & 29.4 & -12.5 \\
\midrule
\multirow{2}{*}{\rotatebox{90}{\scriptsize Closed}} & Gemini-3.1-Pro & 81.1 & 52.1 & 28.8 & -29.0 & 79.0 & 58.9 & 33.1 & -20.1 \\
 & GPT-5.1 & 60.4 & 41.8 & 33.6 & -18.6 & 66.6 & 47.8 & 36.9 & -18.8 \\
\midrule
\multirow{2}{*}{\rotatebox{90}{\scriptsize Hybrid}} & Gemini-3.1-Pro$^\dagger$ & 73.8 & 61.0 & 33.0 & -12.8 & 82.0 & 54.7 & 37.6 & -27.3 \\
 & GPT-5.1$^\dagger$ & 66.2 & 50.6 & 30.9 & -15.5 & 70.4 & 48.6 & 30.5 & -21.9 \\
\midrule
\multirow{1}{*}{\rotatebox{90}{\scriptsize Agt}} & Mobile-Agent-v3 & 48.9 & 34.5 & 30.1 & -14.4 & 34.2 & 27.7 & 30.5 & -6.5 \\
\midrule
\rowcolor{avgrow}
& \textbf{Average} & \textbf{46.0} & \textbf{32.7} & \textbf{29.0} & \textbf{-13.3} & \textbf{45.2} & \textbf{31.4} & \textbf{28.6} & \textbf{-13.8} \\
\bottomrule
\end{tabular}
\end{adjustbox}
\vspace{0.3em}
\begin{minipage}{\textwidth}
\scriptsize
\textit{Notes.} Success rates are percentages, equal-app-weighted; $\Delta$ is in percentage points.
$^\dagger$~\textit{Hybrid}: closed-source planner + UI-Venus-Ground-72B grounder. 
\end{minipage}
\end{table*}

\subsection{RQ2: Performance on Original AndroidWorld Tasks vs. New Tasks}
\label{sec:rq2_template_provenance}

Splitting the benchmark by template provenance shows that the cross-app gap is not an artifact of the newly written AnyAppBench tasks. On the \textbf{AW-inherited templates}, the 13 agents average \textbf{46.0\%} on the original AndroidWorld apps and \textbf{32.7\%} on new apps, a drop of \textbf{13.3} points. On the \textbf{new AnyAppBench templates}, they average \textbf{45.2\%} on the original apps and \textbf{31.4\%} on new apps, a drop of \textbf{13.8} points.

The two groups lose almost the same amount, and their original-app starting points (\textbf{46.0\%} vs.\ \textbf{45.2\%}) are essentially equal. The gap comes from changing the app, not from how the tasks were written: templates taken from AndroidWorld, on which several agents were tuned, transfer no better than templates written for AnyAppBench. Put differently, an evaluation that pairs each template only with its original app overstates same-category generalization by about 13 points for both groups.

\subsection{RQ3: Does App-Independent Sub-Goal Augmentation Close the Cross-App Gap?}
\label{sec:rq3_subgoal_augmentation}

We next ask whether an explicit task decomposition helps agents on new apps. In C2, Gemini-3.1-Pro writes a short list of app-independent sub-goals from the instruction alone. It does not see the screenshot, coordinates, trajectory, or app-specific interface information, and every app of a template receives the same list. The matched comparison covers the five locally served agents on all ten categories with two runs per condition, 5{,}165 paired rollouts in total, one pair per cell and run, excluding pairs whose C2 rollout failed for infrastructure reasons. We report the two category groups separately because the effect differs between them (2{,}285 pairs in SMS, File Manager, Maps, Contacts, and Clock, and 2{,}880 pairs in the remaining five).

Table~\ref{tab:rq3_c2_vs_c1_main} shows a small gain that does not close the gap. Success rises from \textbf{37.8\%} under C1 to \textbf{40.9\%} under C2 (\textbf{+3.2} points, with 265 pairs gained and 193 lost, McNemar $p=0.001$). The gain is larger on the original AndroidWorld apps (\textbf{+6.2} points) than on new apps (\textbf{+2.3} points), so the gap actually \emph{widens}, from 14.9 to 18.8 points. The subgroup
intervals are wide, so we read the split as a direction rather than a precise
size (matched-pair inference in the appendix). The effect also depends on the model and on the run. UI-Venus-7B gains 9.6 points ($p<0.001$), the other four agents change by $+0.7$ to $+2.4$ points without reaching significance, and the second run alone changes by only +2.1 points ($p=0.128$). Sub-goals help an agent carry out a task it can already map onto the original app's interface. They do not supply what it lacks on an unfamiliar one. RQ3 is independent of the human failure attribution in RQ4.

On the remaining five categories (Tasks, Notes, Finance, Music, and Calendar) the effect reverses. Success falls from \textbf{18.3\%} under C1 to \textbf{14.8\%} under C2 (\textbf{$-$3.5} points, 95\% CI $[-5.0, -2.0]$, $p<0.001$), with the largest drops in Calendar ($-$7.3 points) and for Qwen3-VL-8B ($-$10.7 points). Therefore, across all ten categories, instruction-only sub-goals shift success by a few points in either direction depending on the category and the agent, and in neither direction do they close the cross-application gap (see Appendix~\ref{app:rq3_full}).

\begin{table}[t]
\centering
\caption{\textbf{Effect of app-independent sub-goal augmentation}
(2{,}285 matched pairs over the two seeded runs). C1 uses the original
instruction. C2 prepends Gemini-3.1-Pro-generated app-independent
sub-goals. Gap $=$ success on the original AndroidWorld apps minus success
on the new apps. The gap
\emph{widens} under C2; the interaction is a direction, not a precise
size (see the appendix).}
\label{tab:rq3_c2_vs_c1_main}
\scriptsize
\setlength{\tabcolsep}{3pt}
\renewcommand{\arraystretch}{1.15}
\begin{tabularx}{\columnwidth}{@{}l*{4}{>{\centering\arraybackslash}X}@{}}
\toprule
 & \textbf{All} & \textbf{AW app} & \textbf{New} & \textbf{Gap} \\
\midrule
C1 (instruction)   & 37.8 & 49.4 & 34.5 & 14.9 \\
C2 ($+$ sub-goals) & 40.9 & 55.6 & 36.8 & 18.8 \\
\midrule
\rowcolor{avgrow}
$\Delta$ (C2$-$C1) & \textbf{+3.2}$^{*}$ & +6.2 & +2.3 & +3.9 \\
\rowcolor{avgrow}
\tiny 95\% CI & \tiny +0.3, +6.1
 & \tiny $-$2.0, +14.8 & \tiny $-$0.6, +5.2
 & \tiny $-$4.4, +13.1 \\
\bottomrule
\end{tabularx}

\vspace{2pt}
\begin{minipage}{\columnwidth}\scriptsize
$^{*}$ exact McNemar over discordant pairs, $p<0.001$ (265 gained,
193 lost). CIs: agent$\times$app cluster bootstrap, $B{=}10{,}000$.
\end{minipage}
\end{table}
\subsection{RQ4: Application-Driven Shifts in Failure Composition}
\label{sec:rq4_failures}

\begin{figure*}[t]
    \centering
    \includegraphics[width=\textwidth]{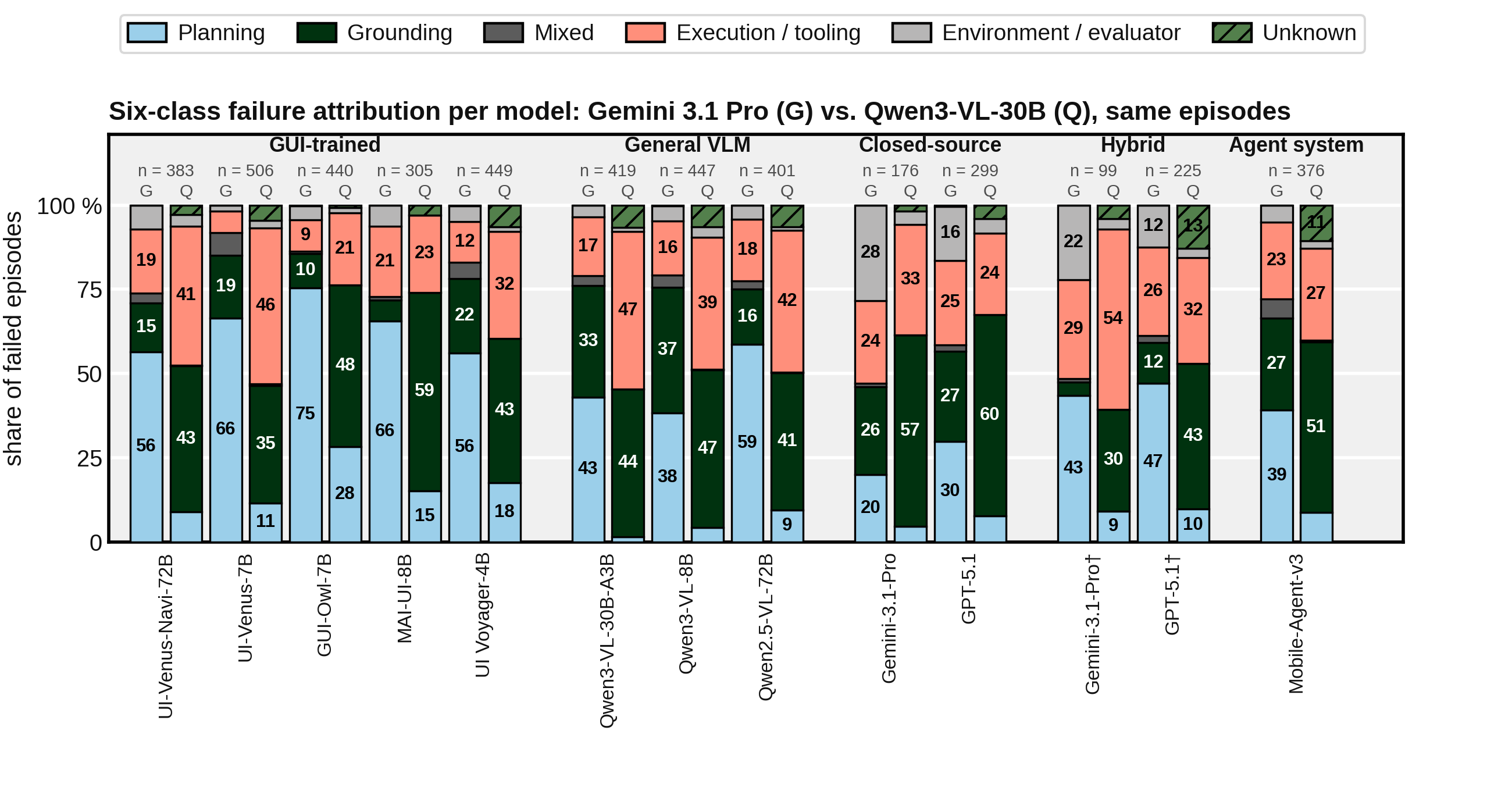}
    \caption{\textbf{Six-class automated failure attribution.} Per-agent
    distributions assigned by Gemini 3.1 Pro (G) and Qwen3-VL-30B-A3B (Q) on
    the same verifier-failed C1 episodes. Bars show the share assigned to each
    failure class; $n$ reports the number of matched episodes for each agent.}
    \label{fig:rq4_failure_bars}
\end{figure*}

Figure~\ref{fig:rq4_failure_bars} shows, for every agent, the six-class
failure composition that each judge assigns to the same verifier-failed
episodes. Two patterns are visible directly in the figure. First, the two
judges disagree systematically: for almost every agent, Gemini attributes most
failures to planning, whereas Qwen attributes them to grounding and
execution/tooling. Second, no single failure class dominates every agent: the mix differs
from agent to agent under both judges, so failure profiles are agent-specific
rather than uniform.

The composition also shifts with the target application. For the primary
reading we group \textit{planning}, \textit{grounding}, and \textit{mixed} as
\emph{planning/grounding-related}, and group \textit{execution/tooling},
\textit{environment/evaluator}, and \textit{unknown} as \emph{non-planning/grounding-related}. Under the
Gemini judge, 79.5\% of the 751 original-app failures are
planning/grounding-related, against 75.2\% of the 3{,}774 new-app failures.
Both judges shift failures toward interface interaction on new apps: the
execution/tooling share rises from 11.3\% to 18.3\% under Gemini and the
grounding share rises from 42.2\% to 46.7\% under Qwen. The exact six-class
split, however, depends on the judge.

Judge--human agreement supports emphasizing the coarse distinction over fine-grained six-class attribution (Table~\ref{tab:human-sampling}). Over the 559 episodes with a completed human label (558 for Qwen, which lacks one label), Gemini matches
the human label on 45.1\% of six-class decisions
($\kappa=0.277$) and on 75.3\% of the binary
planning/grounding-versus-other decisions ($\kappa=0.353$).
Qwen reaches 36.4\% and 68.6\%
($\kappa=0.133$ and $\kappa=0.22$). 

\FloatBarrier
\section{Conclusion}
\label{sec:conclusion}

AnyAppBench isolates within-category cross-application generalization by
holding the user goal and semantic terminal condition fixed while varying the
application. Across 13 agents, success on canonical applications is not a
reliable proxy for success on alternatives supporting the same goal, and
application-independent sub-goals produce only small, category-dependent
changes and do not close the gap. The Gemini diagnostic analysis, validated
against the human reference, shows that the observed composition of failures
depends partly on the target interface; this conclusion rests on the coarse
planning/grounding-related versus non-planning/grounding-related distinction,
while automated judges provide sensitivity analyses. These findings motivate
mobile-agent evaluation over replaceable application sets and continued work
on task decomposition, interface grounding, execution, and recovery under UI
variation.

\section{Limitations}
\label{sec:limitations}

AnyAppBench measures cross-application robustness within a bounded Android
suite. Its 10 categories, 52 applications, and 520 task--application pairs do
not cover the full diversity of platforms, languages, accessibility settings,
or real user workflows. Applications may also differ in permissions,
terminology, workflow length, and internal state representation despite sharing
a semantic success condition. Programmatic verifiers make outcomes auditable,
but app-specific implementations may still miss partially correct or
alternative acceptable solutions.

The human failure analysis is descriptive rather than population-weighted. The
559 episodes are partitioned by category, with one label per
episode and no completed double-annotation set. Therefore, annotator identity is
confounded with category, human--human reliability cannot be estimated, and the
labels are best treated as a human diagnostic reference rather than definitive
ground truth. Fine-grained agreement with this reference also varies across the
Gemini and Qwen judges; our RQ4 conclusion consequently emphasizes the coarser
planning/grounding-related distinction.

AnyAppBench evaluates fixed agents rather than strict unseen-app generalization,
because the training exposure of proprietary and GUI-trained models is generally
unknown. C2 tests only two instruction-only sub-goal generators, and failure
under both C1 and C2 is non-diagnostic because the supplied decomposition may be
incomplete. Finally, live-app evaluation is version-sensitive. We use fixed
APKs and provide version and checksum metadata, but future app updates may alter
interfaces or verifier compatibility. Emulator success also does not certify
safe deployment in messaging, finance, calendar, or contact applications.

\section{Acknowledgments}
This work was partly supported by the Advanced GPU Utilization Support Program and the Institute of Information \& Communications Technology Planning \& Evaluation (IITP), both funded by the Government of the Republic of Korea (Ministry of Science and ICT) (No. RS-2024-00457882, AI Research Hub Project; No. RS-2024-00343989, Enhancing the Ethics of Data Characteristics and Generation AI Models for Social and Ethical Learning; No. RS-2020-II200940, Foundations of Safe Reinforcement Learning and Its Applications to Natural Language Processing; and No. RS-2019-II190075, Artificial Intelligence Graduate School Program (KAIST)).  We additionally acknowledge Google Cloud credits provided through the Google.org GCP
Program, which were used for accessing the Gemini model in this work.

\bibliography{custom}

\clearpage
\appendix

\makeatletter
\@addtoreset{table}{section}
\makeatother
\setcounter{table}{0}
\renewcommand{\thetable}{\thesection.\arabic{table}}
\providecommand{\theHtable}{}
\renewcommand{\theHtable}{appendix.\thesection.\arabic{table}}

\section{Additional Related Work}
\label{app:additional_related}

\begin{table*}[t]
\centering
\footnotesize
\setlength{\tabcolsep}{4pt}
\renewcommand{\arraystretch}{1.16}
\caption{\textbf{Comparison with representative mobile-GUI benchmarks.}
AnyAppBench is the only benchmark shown that satisfies all four criteria.
It contains 520 validated task--app pairs, and a VLM judge diagnoses every
failed trajectory, with human annotations validating the judge
(Appendix~\ref{sec:human-protocol}).}
\label{tab:benchmark-comparison}
\begin{tabularx}{\textwidth}{@{}>{\raggedright\arraybackslash}p{0.225\textwidth}
  *{4}{>{\centering\arraybackslash}X}@{}}
\toprule
\textbf{Benchmark}
& \shortstack{\textbf{Live}\\\textbf{interaction}}
& \shortstack{\textbf{Fixed goal across}\\\textbf{alternative apps}}
& \shortstack{\textbf{Within-category}\\\textbf{app control}}
& \shortstack{\textbf{Trajectory-level}\\\textbf{failure diagnosis}} \\
\midrule
AndroidWorld~\citep{rawles2025androidworld}
& \benchmarkyes & \benchmarkno & \benchmarkno & \benchmarkno \\
AndroidLab~\citep{xu2025androidlab}
& \benchmarkyes & \benchmarkno & \benchmarkno & \benchmarkno \\
SPA-Bench~\citep{chen2025spabench}
& \benchmarkyes & \benchmarkno\textsuperscript{$\dagger$} & \benchmarkno & \benchmarkno \\
GUI-Odyssey~\citep{lu2024guiodyssey}
& \benchmarkno & \benchmarkno\textsuperscript{$\dagger$} & \benchmarkno & \benchmarkno \\
TransBench~\citep{lu2025transbench}
& \benchmarkno & Partial\textsuperscript{$\ddagger$} & \benchmarkno & \benchmarkno \\
\midrule
\textbf{AnyAppBench (ours)}
& \benchmarkyes & \benchmarkyes & \benchmarkyes & \benchmarkyes \\
\bottomrule
\end{tabularx}
\vspace{1pt}
\begin{minipage}{0.99\textwidth}
\scriptsize
\textit{Notes.} \textsuperscript{$\dagger$}Multi-app workflows, not controlled
app substitution; \textsuperscript{$\ddagger$}static cross-app grounding
transfer, not live end-to-end execution.
\end{minipage}
\end{table*}

\paragraph{Mobile GUI benchmarks.} Table~\ref{tab:benchmark-comparison} compares AnyAppBench with representative live, transfer-oriented, and cross-app benchmarks along the evaluation axes most relevant to our setting. More broadly, mobile GUI benchmarks can be divided into offline and online settings. Offline resources such as Android-in-the-Wild~\citep{rawles2023aitw}, AndroidControl~\citep{li2024androidcontrol}, AMEX~\citep{chai2025amex}, Mobile-Bench~\citep{deng2024mobilebench}, Mobile-Bench-v2 \citep{xu2025mobilebenchv2}, MobileVLM/Mobile3M~\citep{wu2024mobilevlm}, ColorBench~\citep{song2025colorbench}, MONDAY~\citep{jang2025monday}, and MobiBench~\citep{im2025mobibench} provide screenshots, UI states, demonstrations, or multi-path trajectory annotations. These benchmarks are scalable and reproducible, but they primarily evaluate fixed states or pre-recorded trajectories. Online benchmarks such as AndroidWorld \citep{rawles2025androidworld}, AndroidLab~\citep{xu2025androidlab}, A3 \citep{chai2025a3}, SPA-Bench~\citep{chen2025spabench}, AndroidArena \citep{xing2024androidarena}, B-MoCA~\citep{lee2024bmoca}, MobileWorld \citep{kong2025mobileworld}, and VenusBench-Mobile~\citep{gong2026venusbench} evaluate agents in live or emulated Android environments. AnyAppBench is complementary to both lines: rather than adding more task types or longer workflows, it holds the functional task fixed and varies the app within the same category.

\paragraph{Cross-app and transfer-oriented evaluation.} Several works study transfer or multi-app behavior. GUI-Odyssey \citep{lu2024guiodyssey} collects trajectories that may move across multiple applications, making it useful for studying cross-app composition. TransBench \citep{lu2025transbench} studies transfer across version, platform, and application axes, but its cross-application setting is primarily framed around static GUI grounding. FedMABench~\citep{wang2025fedmabench} constructs heterogeneous mobile-agent datasets for federated training and shows that app identity can be an important source of heterogeneity. These works motivate AnyAppBench, but their evaluation axes differ from ours. AnyAppBench evaluates the same task template across interchangeable apps in the same functional category, so that performance variation can be attributed to app-specific interface differences rather than to different tasks or app-to-app workflows.

\paragraph{Grounding and diagnostic evaluation.} Element grounding is central to GUI control: the agent must map a textual or visual intent to the correct actionable UI element. SeeClick/ScreenSpot \citep{cheng2024seeclick}, OS-Atlas~\citep{wu2024osatlas}, ScreenSpot-Pro \citep{li2025screenspotpro}, UGround~\citep{gou2025uground}, Set-of-Mark prompting~\citep{yang2023som}, and UI-Ins~\citep{chen2025uiins} study GUI grounding across mobile, desktop, and web interfaces. AnyAppBench connects to this line because its planner-augmented condition gives the acting agent an explicit task decomposition while still requiring it to locate and operate the relevant UI elements in each app. Thus, failures under C2 often expose grounding, interaction, or recovery limitations that are hidden by final success alone.

\paragraph{Evaluation methodology and LLM-based analysis.} As agent benchmarks scale, evaluation design becomes increasingly important. AndroidWorld~\citep{rawles2025androidworld} uses deterministic programmatic success checks, while A3~\citep{chai2025a3}, SPA-Bench \citep{chen2025spabench}, and VenusBench-Mobile \citep{gong2026venusbench} incorporate more flexible automated evaluation for realistic mobile tasks. LlamaTouch~\citep{zhang2024llamatouch} evaluates mobile UI execution by matching essential UI states rather than exact action sequences. In broader agent evaluation, AgentRewardBench \citep{lu2025agentrewardbench}, GUIDE~\citep{zhai2026guide}, and Auto-Eval Judge~\citep{bhonsle2025autoeval} study trajectory-level or judge-based diagnosis. AnyAppBench follows this diagnostic direction, but applies it to within-category cross-app generalization: we compare unaided execution with planner-augmented execution and then use trajectory-level analysis to label planning, grounding, mixed, execution/tooling, environment/evaluator, and unknown failures.

\section{Verifier Design and Coverage}
\label{app:verifier-taxonomy}

Programmatic verifiers alone determine task success. For each retained
task--application pair, the registered task class initializes the episode and
implements \texttt{is\_successful} for the same semantic terminal condition as
the category-level template. The executable inventory resolves all 520 retained
pairs to a registered task class and verifier. Human failure annotation and the
automated judges operate only after this binary decision and never revise it.

\paragraph{Evidence taxonomy.}
A durable-state (\textsc{D}) verifier checks persistent application or Android
system state, such as a database row, stored preference, alarm, contact, or
file. A UI-state (\textsc{U}) verifier checks the final visible or runtime state
when the target property is not exposed durably. A hybrid (\textsc{H}) verifier
combines durable and UI evidence. These labels describe the default semantic
evidence used by a template. Concrete implementations may use app-specific
database paths, content providers, file locations, accessibility queries, or UI
selectors while preserving the same terminal condition.

The accompanying pair-level inventory confirms verifier registration and
coverage for all 520 retained pairs. Table~\ref{tab:verifier_taxonomy} lists
every template with its provenance and evidence type.

\begin{table*}[t]
\centering
\caption{\textbf{Verifier evidence type by task template.}
\textsc{D}: durable-state verifier.
\textsc{U}: UI-state verifier.
\textsc{H}: verifier combining durable and UI evidence or whose evidence source
depends on the application.
Templates marked \textsuperscript{\tiny AW} preserve AndroidWorld task intent
(48 of 100).}
\label{tab:verifier_taxonomy}
\small
\setlength{\tabcolsep}{3pt}
\renewcommand{\arraystretch}{1.08}
\begin{threeparttable}
\begin{tabularx}{\textwidth}{@{}p{0.13\textwidth}X@{}}
\toprule
\textbf{Category} & \textbf{Task templates and verifier evidence type} \\
\midrule
Tasks / To-Do
&
CompletedTasksForDate\textsuperscript{\tiny AW}~(\textsc{U}),
DueNextWeek\textsuperscript{\tiny AW}~(\textsc{U}),
DueOnDate\textsuperscript{\tiny AW}~(\textsc{U}),
HighPriorityTasks\textsuperscript{\tiny AW}~(\textsc{U}),
HighPriorityDueOnDate\textsuperscript{\tiny AW}~(\textsc{U}),
IncompleteTasksOnDate\textsuperscript{\tiny AW}~(\textsc{U}),
DueWithTime~(\textsc{D}),
Recurring~(\textsc{D}),
EditTask~(\textsc{D}),
CompleteTask~(\textsc{D}) \\

\midrule

Notes
&
CreateNote\textsuperscript{\tiny AW}~(\textsc{D}),
EditNote\textsuperscript{\tiny AW}~(\textsc{D}),
MergeNotes\textsuperscript{\tiny AW}~(\textsc{D}),
DeleteNote\textsuperscript{\tiny AW}~(\textsc{D}),
SearchNote\textsuperscript{\tiny AW}~(\textsc{U}),
ShareOrImportNote\textsuperscript{\tiny AW}~(\textsc{H}),
FolderOrMoveNote\textsuperscript{\tiny AW}~(\textsc{D}),
AttachOrTranscribeContent\textsuperscript{\tiny AW}~(\textsc{H}),
CreateChecklist\textsuperscript{\tiny AW}~(\textsc{D}),
CountTodoItems~(\textsc{U}) \\

\midrule

Finance
&
AddExpense\textsuperscript{\tiny AW}~(\textsc{D}),
AddMultipleExpenses\textsuperscript{\tiny AW}~(\textsc{D}),
DeleteTransaction\textsuperscript{\tiny AW}~(\textsc{D}),
EditTransaction\textsuperscript{\tiny AW}~(\textsc{D}),
AddIncome\textsuperscript{\tiny AW}~(\textsc{D}),
DeleteDuplicateTransactions~(\textsc{D}),
AttachReceipt~(\textsc{H}),
CategorySummary~(\textsc{U}),
DateRangeTotal~(\textsc{U}),
TransferBetweenWallets~(\textsc{D}) \\

\midrule

Music
&
CreatePlaylist\textsuperscript{\tiny AW}~(\textsc{D}),
AddToPlaylist\textsuperscript{\tiny AW}~(\textsc{D}),
AddToQueue\textsuperscript{\tiny AW}~(\textsc{U}),
RemoveFromPlaylist\textsuperscript{\tiny AW}~(\textsc{D}),
RenamePlaylist\textsuperscript{\tiny AW}~(\textsc{D}),
SaveOrExportPlaylist~(\textsc{H}),
PlaylistDuration~(\textsc{U}),
ReorderQueue~(\textsc{U}),
SleepTimer~(\textsc{U}),
SearchAndPlay~(\textsc{U}) \\

\midrule

Calendar
&
AddOneEvent\textsuperscript{\tiny AW}~(\textsc{D}),
AddTimedEvent\textsuperscript{\tiny AW}~(\textsc{D}),
AddRepeatingEvent\textsuperscript{\tiny AW}~(\textsc{D}),
DeleteEvent\textsuperscript{\tiny AW}~(\textsc{D}),
EditEvent\textsuperscript{\tiny AW}~(\textsc{D}),
MoveEvent\textsuperscript{\tiny AW}~(\textsc{D}),
EventsOnDate\textsuperscript{\tiny AW}~(\textsc{U}),
NextEvent~(\textsc{U}),
EventsInRange~(\textsc{U}),
AddReminder~(\textsc{D}) \\

\midrule

SMS
&
Send\textsuperscript{\tiny AW}~(\textsc{D}),
Reply\textsuperscript{\tiny AW}~(\textsc{D}),
ReplyMostRecent\textsuperscript{\tiny AW}~(\textsc{D}),
Resend\textsuperscript{\tiny AW}~(\textsc{D}),
SendClipboard\textsuperscript{\tiny AW}~(\textsc{D}),
SendReceivedAddress\textsuperscript{\tiny AW}~(\textsc{D}),
CreateDraftMessage~(\textsc{D}),
EditDraftMessage~(\textsc{D}),
DeleteConversation~(\textsc{D}),
ForwardMessage~(\textsc{D}) \\

\midrule

File Manager
&
DeleteFile\textsuperscript{\tiny AW}~(\textsc{D}),
MoveFile\textsuperscript{\tiny AW}~(\textsc{D}),
SaveCopyOfFile~(\textsc{D}),
CreateFolder~(\textsc{D}),
RenameFile~(\textsc{D}),
ShareFile~(\textsc{U}),
SearchFile~(\textsc{D}),
CompressFiles~(\textsc{D}),
ExtractArchive~(\textsc{D}),
ViewFileInfo~(\textsc{U}) \\

\midrule

Maps
&
AddFavorite\textsuperscript{\tiny AW}~(\textsc{D}),
AddMarker\textsuperscript{\tiny AW}~(\textsc{D}),
RecordTrack\textsuperscript{\tiny AW}~(\textsc{D}),
SearchPlace~(\textsc{U}),
SearchNearbyPlace~(\textsc{U}),
GetDirections~(\textsc{U}),
ShareLocation~(\textsc{U}),
RemoveFavorite~(\textsc{D}),
DeleteMarker~(\textsc{D}),
ExportLocation~(\textsc{H}) \\

\midrule

Contacts
&
AddContact\textsuperscript{\tiny AW}~(\textsc{D}),
NewContactDraft\textsuperscript{\tiny AW}~(\textsc{H}),
EditContact~(\textsc{D}),
DeleteContact~(\textsc{D}),
AddFavoriteContact~(\textsc{D}),
RemoveFavoriteContact~(\textsc{D}),
SearchContact~(\textsc{H}),
ViewContactDetails~(\textsc{H}),
CallContact~(\textsc{H}),
MessageContact~(\textsc{H}) \\

\midrule

Clock
&
CreateTimer\textsuperscript{\tiny AW}~(\textsc{H}),
StopwatchRunning\textsuperscript{\tiny AW}~(\textsc{U}),
PauseStopwatch\textsuperscript{\tiny AW}~(\textsc{U}),
CreateAlarm~(\textsc{H}),
EditAlarm~(\textsc{H}),
EnableAlarm~(\textsc{H}),
DeleteAlarm~(\textsc{H}),
StartTimer~(\textsc{U}),
StopwatchReset~(\textsc{U}),
AddWorldClock~(\textsc{H}) \\

\bottomrule
\end{tabularx}
\begin{tablenotes}[flushleft]
\footnotesize
\item \textsc{D} reads persistent state and \textsc{U} reads the final visible or runtime state. \textsc{H} combines both sources or reads a different source depending on the application (for example, alarm and timer state is stored durably by two Clock applications and only shown on screen by the others). The concrete implementation may use app-specific database paths, content-provider queries, file paths, or accessibility-tree checks while preserving the same terminal condition.
\item For SMS, File Manager, Maps, Contacts, and Clock, every label was verified directly against the success check of the registered task class. For the remaining five categories the labels follow the template inventory.
\end{tablenotes}
\end{threeparttable}
\end{table*}

\paragraph{Scope of the human diagnosis.}
An \textit{environment/evaluator} annotation indicates that the trajectory
contains evidence consistent with an app, emulator, permission, network, or
possible verifier problem. It is a diagnostic hypothesis, not a confirmed
false-positive or false-negative. Confirming verifier correctness requires a
separate audit of the underlying durable and UI evidence; failure-mode
annotation is not such an audit. Therefore, we omit the earlier prospective
verifier-audit subsection and do not report unexecuted audit results.


\FloatBarrier
\section{Human Diagnostic Annotation}
\label{sec:human-protocol}

Programmatic verifiers alone determine whether an evaluation episode succeeds.
For each sampled verifier-failed C1 episode, a human annotator reviews the
trajectory evidence and assigns exactly one of six diagnostic failure-mode
labels. The annotator diagnoses why the trajectory failed but does not revise
the verifier outcome. The resulting human labels provide a reference for
evaluating the reliability of the automated judges and do not affect benchmark
success rates.

\paragraph{Sampling frame.}
The completed reference contains 559 verifier-failed C1 episodes spanning all
ten categories. It includes a fixed-seed uniform sample of 333 episodes from
3,332 failures in Tasks, Notes, Finance, Music, and Calendar. This partition
covers all 13 agents and 29 applications, with 68 original-app and 265 new-app
episodes, all from run~0. The remaining 226 episodes are the labels completed
at submission from a frozen queue of 519 episodes drawn from the 4,529
verifier-failed episodes of the audited three-run block (SMS, File Manager,
Maps, Contacts, and Clock). The queue is a fixed-seed draw stratified so that
every agent contributes at least 10\% of its failures, and it covers all 13
agents, 23 applications, and all three runs. The 226 completed labels comprise
32 original-app and 194 new-app episodes, kept as the latest label per episode.
We sample because annotating the 7,861-episode frozen annotation
sampling frame is impractical at trajectory level. The fixed seeds remove
annotator discretion from episode selection, and the per-agent floor
guarantees that every agent contributes enough failures for the judge--human
comparison instead of letting the most failure-prone agents dominate the
reference.

Each sampled trajectory is reviewed once and assigned exactly one human failure
label. Because there is no completed double-annotation set, human--human
reliability cannot be estimated. Therefore, we use the annotations as a human
validation reference for judge--human agreement rather than as definitive
ground truth.

\begin{table*}[t]
\centering
\footnotesize
\caption{Human-annotation sampling partitions. First-partition counts are taken
from its frozen annotation ledger, and second-partition counts are taken from
the archived submission freeze of the annotation file.}
\label{tab:human-sampling}
\begin{tabularx}{\textwidth}{@{}p{0.13\textwidth}Xp{0.22\textwidth}rp{0.20\textwidth}@{}}
\toprule
\textbf{Partition} & \textbf{Categories} & \textbf{Selection} & \textbf{Episodes} & \textbf{Application role} \\
\midrule
First five & Tasks, Notes, Finance, Music, Calendar & Fixed-seed uniform & 333 & 68 original-app / 265 new-app \\
Remaining five & SMS, File Manager, Maps, Contacts, Clock & Fixed-seed, $\geq$10\% per agent & 226 of 519 & 32 original-app / 194 new-app \\
\midrule
\textbf{Full reference} & \textbf{All ten} & \textbf{Combined} & \textbf{559} & \textbf{100 original-app / 459 new-app} \\
\bottomrule
\end{tabularx}
\end{table*}

\paragraph{Failure taxonomy.}
Each episode receives exactly one label. \textit{Planning} denotes an incorrect
or incomplete task-level strategy, such as selecting the wrong sub-goal,
omitting a required step, or stopping before the intended state is reached.
\textit{Grounding} denotes a reasonable task-level strategy whose intended
action is localized to or executed on the wrong interface element or UI state.
\textit{Mixed} is used when planning and grounding errors both materially affect
the trajectory and neither alone explains the failure. \textit{Execution/tooling}
denotes failure of the action interface or agent framework despite a valid
strategy and target, such as a malformed action, parser error, timeout, or
failed command. \textit{Environment/evaluator} denotes an app, emulator,
permission, or network failure, or evidence suggesting a verifier mismatch.
\textit{Unknown} is used when the available trajectory evidence does not
support reliable attribution.

\begin{figure*}[t]
\centering
\includegraphics[width=\textwidth]{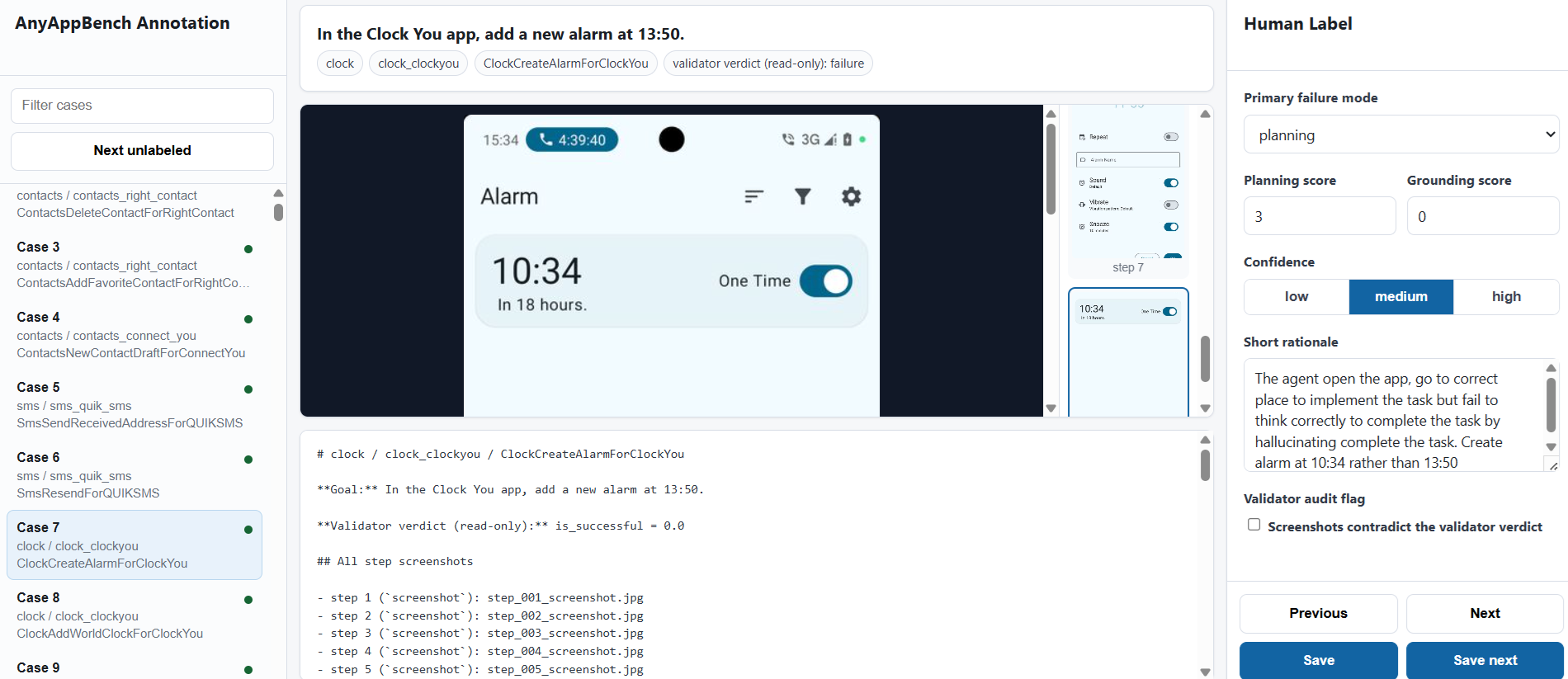}
\caption{\textbf{Human annotation interface.} Web-based tool used by the annotator to review trajectory evidence and assign one diagnostic failure label per episode while all automated-judge outputs are hidden.}
\label{fig:annotation_web}
\end{figure*}

\paragraph{Annotation evidence.}
Figure~\ref{fig:annotation_web} shows the annotation interface. For each sampled episode, the annotator reviews the user goal, task template,
temporally ordered compact action--thought trace, exception information, side
artifacts, salience-selected screenshots, and verifier outcome. The verifier
outcome is presented only as read-only context. Second-partition episodes are
served through a blinded web interface that shows no judge output and records
every label event. After considering the trajectory-level evidence as a whole,
the annotator assigns exactly one label corresponding to the dominant failure
mode. Per-item annotation times are unavailable, so they are not reported.

\begin{table*}[t]
\centering
\small
\caption{Human failure-label counts. Counts describe the annotated sample,
not population failure rates. The combined column concatenates the two
partitions' episode-level labels.}
\label{tab:human-failure-counts}
\begin{tabular}{lrrr}
\toprule
\textbf{Failure label} & \textbf{First five ($N=333$)} & \textbf{Remaining five ($N=226$)} & \textbf{All ($N=559$)} \\
\midrule
Planning              & 50  & 101 & 151 \\
Grounding             & 129 & 65  & 194 \\
Mixed                 & 27  & 24  & 51 \\
Execution/tooling     & 76  & 10  & 86 \\
Environment/evaluator & 45  & 17  & 62 \\
Unknown               & 6   & 9   & 15 \\
\midrule
\textbf{Total}        & \textbf{333} & \textbf{226} & \textbf{559} \\
\bottomrule
\end{tabular}
\end{table*}

\section{VLMs as Diagnostic Judges}
\label{sec:judge-protocol}

Programmatic verifiers alone determine whether an episode succeeds. Separately,
Gemini 3.1 Pro assigns one of six diagnostic failure-mode labels to each
verifier-failed C1 trajectory. Qwen3-VL-30B-A3B-Instruct applies the same
taxonomy as an independent judge-model sensitivity check. Human annotators
label a reference set of 559 verifier-failed C1 episodes under the same
taxonomy (Appendix~\ref{sec:human-protocol}), allowing us to measure agreement
between each automated judge and the human reference. We use the Gemini labels as the diagnostic evidence for RQ4, the human
annotations to measure judge reliability, and the Qwen outputs to assess
sensitivity to the choice of judge model. None of these diagnostic labels changes the verifier outcome or
the benchmark success rates.

\begin{figure*}[t]
  \centering
  \includegraphics[width=\textwidth]{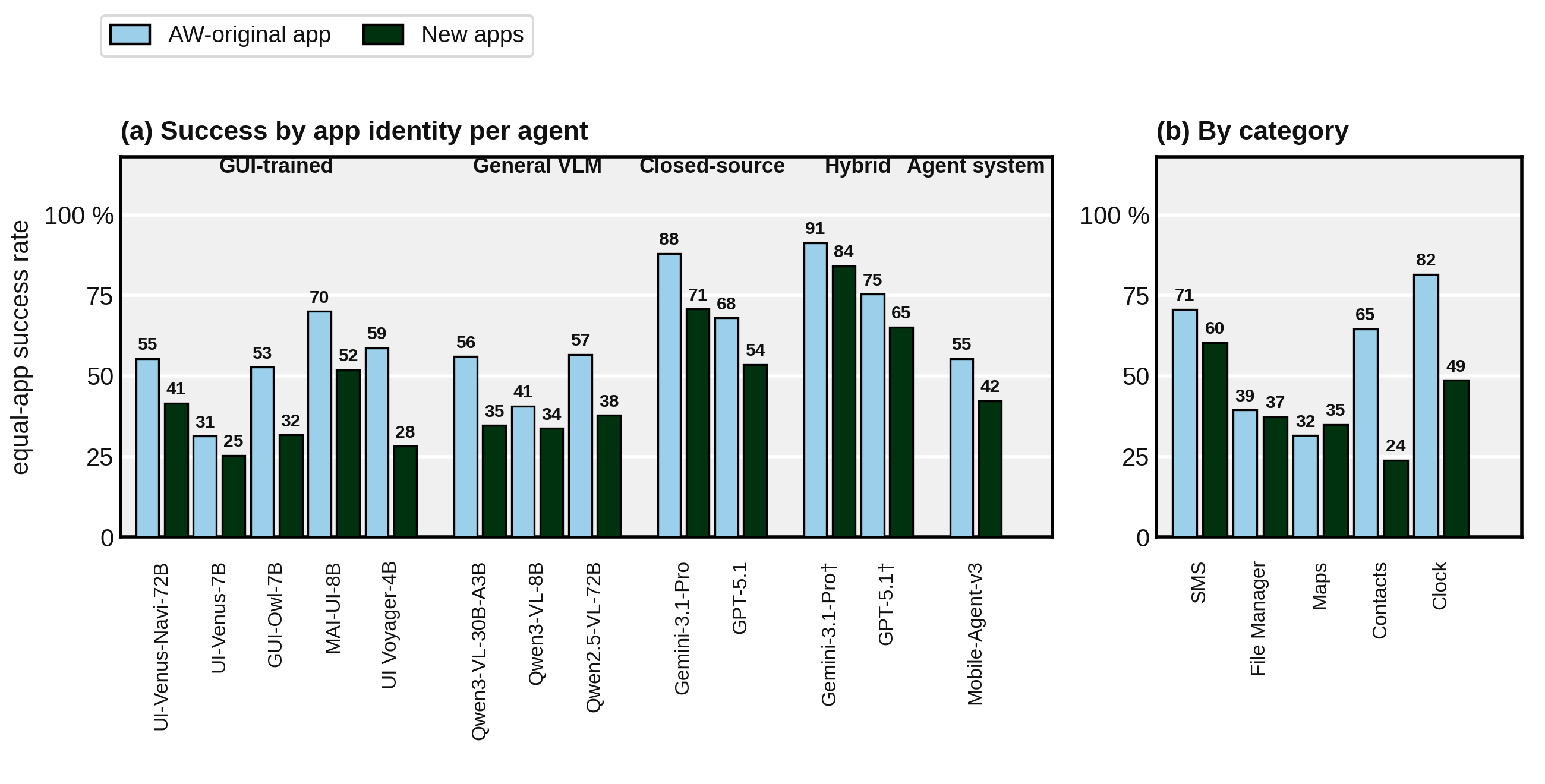}
  \caption{\textbf{Programmatic success context for failure attribution.}
  Equal-app-weighted C1 success on AndroidWorld-original and new applications,
  reported by agent (a) and category (b) for SMS, File Manager, Maps, Contacts,
  and Clock. Programmatic verifiers determine these outcomes. The VLM judges
  are applied only after verification to diagnose failed episodes.}
  \label{fig:judge-cohort-success}
\end{figure*}

\subsection{Models and Decoding}
\label{sec:judge-models}

\begin{table}[t]
\centering
\small
\caption{Automated judges used for sensitivity analysis.}
\label{tab:judge-models}
\begin{tabularx}{\columnwidth}{@{}>{\raggedright\arraybackslash}p{0.44\columnwidth}X@{}}
\toprule
\textbf{Judge} & \textbf{Configuration} \\
\midrule
Gemini 3.1 Pro (\texttt{gemini-3.1-pro-preview}) & Two-stage progressive judge (triage $\rightarrow$ deep review); primary automated sensitivity judge \\
Qwen3-VL-30B-A3B-Instruct (vLLM, 32k context) & Single-call re-judge of the same cases; independent judge-model sensitivity check \\
\bottomrule
\end{tabularx}
\end{table}

\paragraph{Gemini (progressive).} Stage~1 (\emph{triage}) receives the case
payload and only the \textbf{last three} key frames, downscaled to at most
640\,px on the longest edge (JPEG quality~70), under a fast-screening prompt
that must either return a verdict with a confidence level or request
escalation. A case escalates to Stage~2 whenever the triage call sets
\texttt{needs\_escalation}, returns an \texttt{uncertain} verdict, or reports
\texttt{low} confidence. Stage~2 (\emph{deep review}) receives the same payload
and \textbf{all} key frames (at most 896\,px, JPEG quality~80) under a
full-review prompt, and its output replaces the triage output. Both stages use
temperature~0, provider-default \texttt{top\_p}/\texttt{top\_k}, JSON response
mode, a 120-second timeout, and up to three retries with exponential backoff.
The stage that produced the final label, the number of images it saw, and its
token usage are stored with every judgment.

\paragraph{Qwen (re-judge).} Qwen3-VL-30B-A3B-Instruct is served with vLLM
(bf16, 32k-token context, up to eight images per request) and called once per
case with temperature~0 and the same JSON schema. Because the context window
cannot hold every frame of a long trajectory, the re-judge sends at most
\textbf{six} frames chosen by uniform striding over the Gemini key-frame list
(the first and last frames are always kept, and set-of-mark overlays are
preferred when the agent recorded them), again at most 896\,px and JPEG quality~80. If a
request still exceeds the context window it is retried with only the first and
last frame. If the text trace alone overflows (Mobile-Agent-v3 stores
cumulative per-step action histories), the cumulative fields are dropped from
all but the final step, long strings are capped, and only the first three and
last eight steps are kept with an explicit elision marker. Both fallbacks are
recorded in the judgment metadata. The number
of frames each judge saw per case is logged and differs by design (Gemini deep
review sees all frames and Qwen at most six), so the two judges are
\emph{not} frame-matched. This is reported as a limitation of the cross-judge
comparison.

\paragraph{Estimated API cost.}
The GPT-5.1 and API-backed hybrid evaluations with $K=3$ cost approximately
\$250. For Gemini judging, recorded calls average 9,238 input and 230 output
tokens. Extrapolated to the 4,529 audited-block failures, this corresponds to
approximately 41.8 million input tokens (\$84) and 1.0 million output tokens
(\$13), or roughly \$97 in total at standard Gemini 3.1 Pro Preview pricing.
\footnote{Gemini Developer API pricing:
\url{https://ai.google.dev/gemini-api/docs/pricing}.}
The resulting API subtotal is approximately \$347. This is a lower bound
because the two-stage judge issues extra triage calls and because Gemini
acting-agent runs and C2 sub-goal generation are additional but lack itemized
usage logs. Qwen and the remaining open models were locally served, so local
compute costs are excluded.

\subsection{Key-Frame Selection}
\label{sec:keyframe}

Key frames are selected per episode from the stored screenshots
(\texttt{--smart\_steps}). The selector always includes the first and last
screenshots, then adds error-adjacent steps, the onset of repeated-action
loops, and transitions between high-level action types. With no frame cap
(the Gemini deep stage) this yields every step of the trajectory in order,
and with a cap of $k$ frames (six for the Qwen re-judge, the last three for
the triage stage) the remaining slots are filled by uniform striding over the
selected list. For Mobile-Agent-v3, whose recorder stores a three-frame sliding
window per step, the most recent frame of each window is used.

\subsection{Classifier Prompt}
\label{sec:judge-prompt}

All prompts share the same six-class output schema and definitions
(\textit{planning}, \textit{grounding}, \textit{mixed}, \textit{execution/tooling},
\textit{environment/evaluator}, \textit{unknown}). The system prompts instruct
the model not to revise \texttt{is\_successful=0}, to return one JSON object,
and to cite trajectory or screenshot evidence for every non-trivial diagnosis.
The triage prompt additionally requires escalation whenever the evidence is
incomplete. An \texttt{environment\_or\_evaluator} output records a suspected
harness or verifier problem rather than a confirmed one. Every such case in the
five-category matrix was inspected by hand, and the cases that turned out to be
infrastructure faults (stale phone calls, failed seeding, emulator faults) were
re-run before any number in this paper was computed.

\begin{promptbox}{Failure-mode classifier system prompt}
\begin{lstlisting}[style=promptstyle]
You are an AnyAppBench failure-mode classifier.

You receive an AnyAppBench episode whose programmatic validator already returned
is_successful = 0. Your job is NOT to second-guess that verdict. Your job is
to classify the dominant reason the agent failed.

You see: the goal, the recorded `is_successful` flag (always 0 for cases
routed to you), a compact step trace, side artifacts, and a STITCHED set of
key screenshots (first, last, error-adjacent, repeated-action, and any
transition steps).

Return one JSON object only, exactly:
{
  "primary_failure_mode": "planning|grounding|mixed_planning_grounding|execution_tooling|environment_or_evaluator|unknown",
  "planning_score": 0-3,
  "grounding_score": 0-3,
  "confidence": "low|medium|high",
  "rationale": "one concise paragraph",
  "evidence": ["short quoted/paraphrased evidence", "screenshot:step_4 shows ..."]
}

Rules:
- "planning" applies when the agent chose a wrong subgoal, missed a required
  step, stopped early, or committed to an infeasible strategy.
- "grounding" applies when the plan was right but the agent could not locate
  or interact with the correct UI element.
- "mixed_planning_grounding" applies when both meaningfully contributed.
- "execution_tooling" applies when the agent's framework (parser, tool call,
  API timeout, malformed action) is the dominant failure source.
- "environment_or_evaluator" applies when an app crash, missing permission,
  network outage, or a likely validator false-negative is the dominant
  source. A true validator false-negative is reported separately by the
  validator audit in Sec. 6.7; flagging one here is a hypothesis, not a
  verdict.
- "unknown" when the evidence is genuinely insufficient.
- Scores: 0 absent / 1 possible / 2 likely / 3 dominant for each axis.
- Cite at least one step number or screenshot index per non-trivial mode.
- Do NOT include markdown, prose, or extra fields.
\end{lstlisting}
\end{promptbox}

The prompt is reproduced verbatim from the archived production configuration. Its reference to ``Sec.~6.7'' points to an internal draft section and does not correspond to a section in the present manuscript.

\subsection{Judge--Human Agreement}
\label{sec:judge-validation}

\makeatletter
\setlength{\@dblfptop}{0pt}
\setlength{\@dblfpsep}{10pt plus 2pt minus 2pt}
\setlength{\@dblfpbot}{0pt plus 1fil}
\makeatother
\begin{table*}[t]
\centering
\small
\caption{Agreement between the automated judges and the human reference. Block A is the 333-label first-five partition and Block B the 226-label second-five partition. Pooled $n$ is 559 for Gemini and 558 for Qwen, which lacks one label. Gemini's coarse agreement is similar across blocks (76.0 versus 74.3).}
\label{tab:judge-human-agreement}
\textbf{(a) Overall agreement}\\[-2pt]
\begin{tabular}{llrrr}
\toprule
\textbf{Judge} & \textbf{Metric} & \textbf{Block A} & \textbf{Block B} & \textbf{Pooled} \\
\midrule
Gemini-3.1-Pro & six-class agreement (\%) & 48.9 & 39.4 & 45.1 \\
 & six-class $\kappa$ & 0.351 & 0.103 & 0.277 \\
 & coarse agreement (\%) & 76.0 & 74.3 & 75.3 \\
 & coarse $\kappa$ & 0.447 & 0.139 & 0.353 \\
\midrule
Qwen3-VL-30B & six-class agreement (\%) & 43.8 & 25.3 & 36.4 \\
 & six-class $\kappa$ & 0.260 & 0.053 & 0.133 \\
 & coarse agreement (\%) & 75.7 & 58.2 & 68.6 \\
 & coarse $\kappa$ & 0.428 & 0.010 & 0.220 \\
\bottomrule
\end{tabular}

\vspace{5pt}
\textbf{(b) Per-class F$_1$}\\[-2pt]
\label{tab:judge-per-class-f1}
\begin{tabularx}{\textwidth}{@{}l*{6}{>{\centering\arraybackslash}X}@{}}
\toprule
\textbf{Judge}
& \textbf{Planning}
& \textbf{Grounding}
& \textbf{Mixed}
& \shortstack{\textbf{Execution/}\\\textbf{tooling}}
& \shortstack{\textbf{Environment/}\\\textbf{evaluator}}
& \textbf{Unknown} \\
\midrule
Gemini & 0.42 & 0.59 & 0.18 & 0.66 & 0.44 & 0.17 \\
Qwen   & 0.34 & 0.50 & 0.00 & 0.64 & 0.17 & 0.29 \\
\bottomrule
\end{tabularx}
\end{table*}

The PG-related binary scheme groups planning, grounding, and mixed, and groups
execution/tooling, environment/evaluator, and unknown as
non-planning/grounding-related. Although agreement mechanically increases when
labels are collapsed, both judges agree with the human reference more often on
this broad distinction than on six-class attribution. Human annotation serves
as the external reference for judging label reliability.
The two automated judges also disagree with each other. On the 4{,}525
audited-block episodes judged by both, cross-judge six-class agreement is
25.9\% (95\% cluster bootstrap CI $[24.3, 27.5]$) with Cohen's $\kappa$ of
0.056, and the coarse split reaches 56.9\% ($\kappa=0.07$). This cross-judge
analysis is distinct from judge--human agreement and is not used as evidence
of human validity.

\FloatBarrier
\section{Per-Category, Per-App Success Rates}
\label{app:full_results}

In this appendix we report success rates broken down by individual app, for each of the ten functional categories in AnyAppBench. The structure is identical across all ten tables. Each app has two sub-columns: \textbf{AW} reports the success rate on AW-inherited templates evaluated on that app, and \textbf{New} reports the success rate on the newly added AnyAppBench templates. This decomposition follows the same template-provenance split used in Table~\ref{tab:main_cross_app_gap_recomputed} of the main paper, applied here at the per-app granularity. The leftmost shaded sub-columns belong to the AndroidWorld-original baseline app(s) for each category and the remaining sub-columns belong to the newly installed apps. The bottom row of each table reports the column-wise mean across model rows with completed results, exposing per-app difficulty independent of model choice. Aggregate summaries (New Avg., Std., $\Delta$) are reported in the main paper and are not duplicated here.

%
\newcolumntype{Y}{>{\columncolor{awcol}\centering\arraybackslash}X}
\newcolumntype{Z}{>{\columncolor{catcol}\centering\arraybackslash}X}

\newcommand{\AppFamilyLabel}[1]{%
  \makecell[l]{\footnotesize #1}%
}

\newcommand{\AppTableSetup}{%
  \footnotesize
  \setlength{\tabcolsep}{1.6pt}%
  \renewcommand{\arraystretch}{1.04}%
}
\newcommand{\AppLevelNoteRuns}{%
  \vspace{1pt}%
  \begin{tablenotes}[para,flushleft]
    \footnotesize\raggedright
    \item[] \textit{Note.} Cells show success rates (\%) pooling the three runs of every template.
    \textbf{AW}: AW-inherited templates.
    \textbf{New}: newly added AnyAppBench templates.
    $^\dagger$~Planner--executor setup: the closed-source model plans and
    UI-Venus-72B grounds.
    The bottom row reports the column-wise mean success rate over the 13 model rows.
  \end{tablenotes}%
}

\newcommand{\AppLevelCaption}[1]{\caption{#1}}

\newcommand{\AppLevelNote}{%
  \vspace{1pt}%
  \begin{tablenotes}[para,flushleft]
    \footnotesize\raggedright
    \item[] \textit{Note.} Cells show successful tasks out of scheduled templates.
    \textbf{AW}: AW-inherited templates.
    \textbf{New}: newly added AnyAppBench templates.
    $^\dagger$~Planner--executor setup: the closed-source model plans and
    UI-Venus-72B grounds.
    The bottom row reports the column-wise mean success rate over rows with
    completed results.
  \end{tablenotes}%
}
\begin{table*}[h]
\centering
\begin{threeparttable}
\AppTableSetup

\begin{tabularx}{\textwidth}{@{}ll*{6}{YZ}@{}}
\toprule
& & \multicolumn{2}{c}{\cellcolor{awshade}\textbf{Tasks.org}} & \multicolumn{2}{c}{\textbf{Cfait}} & \multicolumn{2}{c}{\makecell{\textbf{Todo List}\\\textbf{(PFA)}}} & \multicolumn{2}{c}{\textbf{ntodotxt}} & \multicolumn{2}{c}{\textbf{TaskMate}} & \multicolumn{2}{c}{\textbf{Grit}} \\
\cmidrule(lr){3-4}\cmidrule(lr){5-6}\cmidrule(lr){7-8}\cmidrule(lr){9-10}\cmidrule(lr){11-12}\cmidrule(lr){13-14}
\textbf{Family} & \textbf{Model} & \textbf{AW} & \textbf{New} & \textbf{AW} & \textbf{New} & \textbf{AW} & \textbf{New} & \textbf{AW} & \textbf{New} & \textbf{AW} & \textbf{New} & \textbf{AW} & \textbf{New} \\
\midrule
\multirow{5}{*}{\AppFamilyLabel{GUI\\Trained}} & UI-Venus-Navi-72B & 27.8 & 8.3 & 38.9 & 8.3 & 22.2 & 8.3 & 22.2 & 0.0 & 22.2 & 0.0 & 22.2 & 0.0 \\
 & UI-Venus-7B & 77.8 & 33.3 & 22.2 & 0.0 & 22.2 & 8.3 & 38.9 & 0.0 & 55.6 & 33.3 & 22.2 & 8.3 \\
 & GUI-Owl-7B & 61.1 & 33.3 & 22.2 & 0.0 & 22.2 & 8.3 & 22.2 & 0.0 & 55.6 & 33.3 & 22.2 & 8.3 \\
 & MAI-UI-8B & 94.4 & 75.0 & 72.2 & 0.0 & 22.2 & 8.3 & 22.2 & 8.3 & 55.6 & 8.3 & 88.9 & 33.3 \\
 & UI Voyager-4B & 77.8 & 0.0 & 38.9 & 8.3 & 55.6 & 8.3 & 88.9 & 0.0 & 72.2 & 0.0 & 88.9 & 8.3 \\
\midrule
\multirow{3}{*}{\AppFamilyLabel{General}} & Qwen3-VL-30B-A3B & 44.4 & 0.0 & 22.2 & 0.0 & 22.2 & 0.0 & 38.9 & 0.0 & 22.2 & 0.0 & 22.2 & 0.0 \\
 & Qwen3-VL-8B & 27.8 & 0.0 & 22.2 & 0.0 & 22.2 & 0.0 & 38.9 & 0.0 & 22.2 & 0.0 & 22.2 & 8.3 \\
 & Qwen2.5-VL-72B & 27.8 & 0.0 & 22.2 & 0.0 & 22.2 & 8.3 & 22.2 & 0.0 & 22.2 & 0.0 & 22.2 & 0.0 \\
\midrule
\multirow{2}{*}{\AppFamilyLabel{Closed}} & Gemini-3.1-Pro & 77.8 & 75.0 & 72.2 & 33.3 & 33.3 & 25.0 & 38.9 & 8.3 & 72.2 & 0.0 & 50.0 & 41.7 \\
 & GPT-5.1 & 27.8 & 25.0 & 38.9 & 0.0 & 72.2 & 33.3 & 38.9 & 8.3 & 38.9 & 8.3 & 55.6 & 33.3 \\
\midrule
\multirow{2}{*}{\AppFamilyLabel{Hybrid}} & Gemini-3.1-Pro$^\dagger$ & 38.9 & 41.7 & 55.6 & 8.3 & 27.8 & 25.0 & 55.6 & 8.3 & 38.9 & 0.0 & 50.0 & 41.7 \\
 & GPT-5.1$^\dagger$ & 27.8 & 25.0 & 94.4 & 25.0 & 33.3 & 25.0 & 88.9 & 8.3 & 22.2 & 0.0 & 50.0 & 41.7 \\
\midrule
\AppFamilyLabel{Agent} & Mobile-Agent-v3 & 27.8 & 25.0 & 22.2 & 0.0 & 22.2 & 0.0 & 22.2 & 0.0 & 22.2 & 0.0 & 44.4 & 8.3 \\
\midrule
\rowcolor{appavgrow}
\multicolumn{2}{l}{\textbf{Mean success rate}} & 49.2 & 26.3 & 41.9 & 6.4 & 30.8 & 12.2 & 41.5 & 3.2 & 40.2 & 6.4 & 43.2 & 17.9 \\
\bottomrule
\end{tabularx}
\AppLevelCaption{\textbf{App-level success rates for Tasks / To-Do (\%, three runs pooled).} Each app has two sub-columns: AW-inherited templates (6) and newly added AnyAppBench templates (4), with three runs per template. Each cell averages 18 or 12 rollouts.}
\label{tab:app_level_tasks}
\AppLevelNoteRuns
\end{threeparttable}
\end{table*}

\begin{table*}[t]
\centering
\begin{threeparttable}
\AppTableSetup
\renewcommand{\arraystretch}{1.08}

\begin{tabularx}{\textwidth}{@{}ll*{6}{YZ}@{}}
\toprule
\multirow{2}{*}{\textbf{Family}} & \multirow{2}{*}{\textbf{Model}}
& \multicolumn{2}{c}{\cellcolor{awshade}\makecell{\textbf{Joplin}}}
& \multicolumn{2}{c}{\cellcolor{awshade}\makecell{\textbf{Markor}}}
& \multicolumn{2}{c}{\makecell{\textbf{NotallyX}}}
& \multicolumn{2}{c}{\makecell{\textbf{neutriNote}\\[-1pt]\textbf{CE}}}
& \multicolumn{2}{c}{\makecell{\textbf{Notesnook}}}
& \multicolumn{2}{c}{\makecell{\textbf{Orgzly}\\[-1pt]\textbf{Revived}}} \\
\cmidrule(lr){3-4}\cmidrule(lr){5-6}\cmidrule(lr){7-8}\cmidrule(lr){9-10}\cmidrule(lr){11-12}\cmidrule(lr){13-14}
& & \textbf{AW} & \textbf{New} & \textbf{AW} & \textbf{New} & \textbf{AW} & \textbf{New} & \textbf{AW} & \textbf{New} & \textbf{AW} & \textbf{New} & \textbf{AW} & \textbf{New} \\
\midrule
\multirow{5}{*}{\AppFamilyLabel{GUI\\Trained}} & UI-Venus-Navi-72B & 29.6 & 0.0 & 0.0 & 0.0 & 11.1 & 0.0 & 3.7 & 33.3 & 7.4 & 0.0 & 7.4 & 0.0 \\
 & UI-Venus-7B & 40.7 & 0.0 & 11.1 & 0.0 & 29.6 & 0.0 & 18.5 & 0.0 & 40.7 & 33.3 & 40.7 & 0.0 \\
 & GUI-Owl-7B & 40.7 & 0.0 & 3.7 & 0.0 & 59.3 & 33.3 & 40.7 & 33.3 & 18.5 & 33.3 & 44.4 & 33.3 \\
 & MAI-UI-8B & 77.8 & 66.7 & 59.3 & 0.0 & 51.9 & 33.3 & 3.7 & 33.3 & 51.9 & 0.0 & 44.4 & 0.0 \\
 & UI Voyager-4B & 25.9 & 0.0 & 3.7 & 0.0 & 59.3 & 0.0 & 51.9 & 0.0 & 11.1 & 0.0 & 33.3 & 33.3 \\
\midrule
\multirow{3}{*}{\AppFamilyLabel{General}} & Qwen3-VL-30B-A3B & 14.8 & 0.0 & 3.7 & 0.0 & 7.4 & 0.0 & 3.7 & 0.0 & 0.0 & 0.0 & 0.0 & 0.0 \\
 & Qwen3-VL-8B & 3.7 & 0.0 & 3.7 & 0.0 & 3.7 & 0.0 & 0.0 & 0.0 & 3.7 & 0.0 & 0.0 & 0.0 \\
 & Qwen2.5-VL-72B & 51.9 & 0.0 & 3.7 & 0.0 & 18.5 & 33.3 & 7.4 & 33.3 & 11.1 & 0.0 & 22.2 & 33.3 \\
\midrule
\multirow{2}{*}{\AppFamilyLabel{Closed}} & Gemini-3.1-Pro & 88.9 & 66.7 & 70.4 & 66.7 & 59.3 & 33.3 & 7.4 & 33.3 & 40.7 & 33.3 & 44.4 & 33.3 \\
 & GPT-5.1 & 88.9 & 33.3 & 70.4 & 0.0 & 40.7 & 33.3 & 22.2 & 33.3 & 29.6 & 33.3 & 44.4 & 0.0 \\
\midrule
\multirow{2}{*}{\AppFamilyLabel{Hybrid}} & Gemini-3.1-Pro$^\dagger$ & 88.9 & 33.3 & 70.4 & 33.3 & 29.6 & 0.0 & 7.4 & 0.0 & 51.9 & 0.0 & 55.6 & 0.0 \\
 & GPT-5.1$^\dagger$ & 88.9 & 33.3 & 70.4 & 33.3 & 51.9 & 33.3 & 33.3 & 0.0 & 33.3 & 0.0 & 11.1 & 0.0 \\
\midrule
\AppFamilyLabel{Agent} & Mobile-Agent-v3 & 37.0 & 33.3 & 3.7 & 0.0 & 40.7 & 0.0 & 33.3 & 0.0 & 11.1 & 0.0 & 11.1 & 0.0 \\
\midrule
\rowcolor{appavgrow}
\multicolumn{2}{l}{\textbf{Mean success rate}} & 52.1 & 20.5 & 28.8 & 10.3 & 35.6 & 15.4 & 17.9 & 15.4 & 23.9 & 10.2 & 27.6 & 10.2 \\
\bottomrule
\end{tabularx}
\AppLevelCaption{\textbf{App-level success rates for Notes (\%, three runs pooled).} Each app has two sub-columns: AW-inherited templates (9) and newly added AnyAppBench templates (1), with three runs per template. Each cell averages 27 or 3 rollouts.}
\label{tab:app_level_notes}
\AppLevelNoteRuns
\end{threeparttable}
\end{table*}

\begin{table*}[t]
\centering
\begin{threeparttable}
\AppTableSetup

\begin{tabularx}{\textwidth}{@{}ll*{6}{YZ}@{}}
\toprule
& & \multicolumn{2}{c}{\cellcolor{awshade}\textbf{Pro Expense}} & \multicolumn{2}{c}{\textbf{Oinkoin}} & \multicolumn{2}{c}{\makecell{\textbf{OpenMoney}\\\textbf{Box}}} & \multicolumn{2}{c}{\textbf{My Expenses}} & \multicolumn{2}{c}{\makecell{\textbf{Finance}\\\textbf{Manager}}} & \multicolumn{2}{c}{\textbf{Sushi}} \\
\cmidrule(lr){3-4}\cmidrule(lr){5-6}\cmidrule(lr){7-8}\cmidrule(lr){9-10}\cmidrule(lr){11-12}\cmidrule(lr){13-14}
\textbf{Family} & \textbf{Model} & \textbf{AW} & \textbf{New} & \textbf{AW} & \textbf{New} & \textbf{AW} & \textbf{New} & \textbf{AW} & \textbf{New} & \textbf{AW} & \textbf{New} & \textbf{AW} & \textbf{New} \\
\midrule
\multirow{5}{*}{\AppFamilyLabel{GUI\\Trained}} & UI-Venus-Navi-72B & 33.3 & 0.0 & 20.0 & 13.3 & 6.7 & 0.0 & 26.7 & 0.0 & 6.7 & 6.7 & 6.7 & 0.0 \\
 & UI-Venus-7B & 33.3 & 0.0 & 26.7 & 6.7 & 26.7 & 0.0 & 26.7 & 20.0 & 6.7 & 6.7 & 46.7 & 6.7 \\
 & GUI-Owl-7B & 13.3 & 0.0 & 20.0 & 13.3 & 26.7 & 6.7 & 6.7 & 6.7 & 20.0 & 13.3 & 20.0 & 13.3 \\
 & MAI-UI-8B & 93.3 & 13.3 & 60.0 & 13.3 & 46.7 & 6.7 & 53.3 & 6.7 & 66.7 & 0.0 & 60.0 & 13.3 \\
 & UI Voyager-4B & 93.3 & 33.3 & 6.7 & 6.7 & 46.7 & 0.0 & 53.3 & 0.0 & 46.7 & 6.7 & 26.7 & 0.0 \\
\midrule
\multirow{3}{*}{\AppFamilyLabel{General}} & Qwen3-VL-30B-A3B & 13.3 & 0.0 & 6.7 & 0.0 & 6.7 & 0.0 & 13.3 & 0.0 & 13.3 & 0.0 & 6.7 & 0.0 \\
 & Qwen3-VL-8B & 13.3 & 0.0 & 26.7 & 6.7 & 6.7 & 0.0 & 13.3 & 0.0 & 13.3 & 0.0 & 13.3 & 6.7 \\
 & Qwen2.5-VL-72B & 33.3 & 13.3 & 26.7 & 6.7 & 6.7 & 0.0 & 33.3 & 6.7 & 33.3 & 6.7 & 13.3 & 6.7 \\
\midrule
\multirow{2}{*}{\AppFamilyLabel{Closed}} & Gemini-3.1-Pro & 93.3 & 13.3 & 26.7 & 0.0 & 26.7 & 6.7 & 80.0 & 13.3 & 40.0 & 26.7 & 40.0 & 13.3 \\
 & GPT-5.1 & 100.0 & 13.3 & 66.7 & 6.7 & 73.3 & 0.0 & 80.0 & 6.7 & 73.3 & 6.7 & 73.3 & 0.0 \\
\midrule
\multirow{2}{*}{\AppFamilyLabel{Hybrid}} & Gemini-3.1-Pro$^\dagger$ & 73.3 & 6.7 & 26.7 & 0.0 & 13.3 & 6.7 & 80.0 & 13.3 & 73.3 & 6.7 & 20.0 & 13.3 \\
 & GPT-5.1$^\dagger$ & 73.3 & 6.7 & 40.0 & 13.3 & 13.3 & 6.7 & 40.0 & 13.3 & 60.0 & 13.3 & 60.0 & 13.3 \\
\midrule
\AppFamilyLabel{Agent} & Mobile-Agent-v3 & 73.3 & 6.7 & 53.3 & 6.7 & 33.3 & 0.0 & 33.3 & 6.7 & 40.0 & 33.3 & 53.3 & 6.7 \\
\midrule
\rowcolor{appavgrow}
\multicolumn{2}{l}{\textbf{Mean success rate}} & 56.9 & 8.2 & 31.3 & 7.2 & 25.7 & 2.6 & 41.5 & 7.2 & 37.9 & 9.8 & 33.8 & 7.2 \\
\bottomrule
\end{tabularx}
\AppLevelCaption{\textbf{App-level success rates for Finance (\%, three runs pooled).} Each app has two sub-columns: AW-inherited templates (5) and newly added AnyAppBench templates (5), with three runs per template. Each cell averages 15 rollouts.}
\label{tab:app_level_finance}
\AppLevelNoteRuns
\end{threeparttable}
\end{table*}

\begin{table*}[t]
\centering
\begin{threeparttable}
\AppTableSetup

\begin{tabularx}{\textwidth}{@{}ll*{6}{YZ}@{}}
\toprule
& & \multicolumn{2}{c}{\cellcolor{awshade}\textbf{Retro Music}} & \multicolumn{2}{c}{\makecell{\textbf{Fossify Music}\\\textbf{Player}}} & \multicolumn{2}{c}{\textbf{Apollo}} & \multicolumn{2}{c}{\textbf{SicMu Neo}} & \multicolumn{2}{c}{\makecell{\textbf{Phonograph}\\\textbf{Plus}}} & \multicolumn{2}{c}{\makecell{\textbf{Monster}\\\textbf{Music}}} \\
\cmidrule(lr){3-4}\cmidrule(lr){5-6}\cmidrule(lr){7-8}\cmidrule(lr){9-10}\cmidrule(lr){11-12}\cmidrule(lr){13-14}
\textbf{Family} & \textbf{Model} & \textbf{AW} & \textbf{New} & \textbf{AW} & \textbf{New} & \textbf{AW} & \textbf{New} & \textbf{AW} & \textbf{New} & \textbf{AW} & \textbf{New} & \textbf{AW} & \textbf{New} \\
\midrule
\multirow{5}{*}{\AppFamilyLabel{GUI\\Trained}} & UI-Venus-Navi-72B & 46.7 & 20.0 & 13.3 & 6.7 & 13.3 & 0.0 & 20.0 & 0.0 & 20.0 & 0.0 & 40.0 & 0.0 \\
 & UI-Venus-7B & 26.7 & 20.0 & 13.3 & 6.7 & 13.3 & 0.0 & 20.0 & 0.0 & 20.0 & 0.0 & 40.0 & 0.0 \\
 & GUI-Owl-7B & 26.7 & 13.3 & 13.3 & 0.0 & 13.3 & 0.0 & 20.0 & 0.0 & 20.0 & 0.0 & 20.0 & 0.0 \\
 & MAI-UI-8B & 26.7 & 13.3 & 33.3 & 0.0 & 20.0 & 0.0 & 20.0 & 6.7 & 20.0 & 0.0 & 40.0 & 6.7 \\
 & UI Voyager-4B & 26.7 & 20.0 & 13.3 & 0.0 & 20.0 & 0.0 & 20.0 & 6.7 & 20.0 & 0.0 & 20.0 & 0.0 \\
\midrule
\multirow{3}{*}{\AppFamilyLabel{General}} & Qwen3-VL-30B-A3B & 26.7 & 13.3 & 33.3 & 0.0 & 20.0 & 0.0 & 20.0 & 0.0 & 20.0 & 0.0 & 20.0 & 0.0 \\
 & Qwen3-VL-8B & 26.7 & 13.3 & 13.3 & 0.0 & 20.0 & 0.0 & 20.0 & 0.0 & 20.0 & 0.0 & 20.0 & 6.7 \\
 & Qwen2.5-VL-72B & 66.7 & 13.3 & 13.3 & 6.7 & 20.0 & 6.7 & 20.0 & 0.0 & 20.0 & 0.0 & 60.0 & 0.0 \\
\midrule
\multirow{2}{*}{\AppFamilyLabel{Closed}} & Gemini-3.1-Pro & 86.7 & 26.7 & 73.3 & 40.0 & 20.0 & 6.7 & 20.0 & 13.3 & 73.3 & 20.0 & 93.3 & 40.0 \\
 & GPT-5.1 & 46.7 & 26.7 & 53.3 & 20.0 & 20.0 & 6.7 & 20.0 & 13.3 & 53.3 & 20.0 & 40.0 & 33.3 \\
\midrule
\multirow{2}{*}{\AppFamilyLabel{Hybrid}} & Gemini-3.1-Pro$^\dagger$ & 93.3 & 26.7 & 93.3 & 40.0 & 40.0 & 6.7 & 20.0 & 6.7 & 80.0 & 6.7 & 93.3 & 40.0 \\
 & GPT-5.1$^\dagger$ & 93.3 & 26.7 & 93.3 & 40.0 & 40.0 & 6.7 & 20.0 & 6.7 & 80.0 & 0.0 & 73.3 & 40.0 \\
\midrule
\AppFamilyLabel{Agent} & Mobile-Agent-v3 & 46.7 & 13.3 & 20.0 & 6.7 & 20.0 & 0.0 & 20.0 & 0.0 & 20.0 & 0.0 & 40.0 & 0.0 \\
\midrule
\rowcolor{appavgrow}
\multicolumn{2}{l}{\textbf{Mean success rate}} & 49.3 & 19.0 & 36.9 & 12.8 & 21.5 & 2.6 & 20.0 & 4.1 & 35.9 & 3.6 & 46.1 & 12.8 \\
\bottomrule
\end{tabularx}
\AppLevelCaption{\textbf{App-level success rates for Music (\%, three runs pooled).} Each app has two sub-columns: AW-inherited templates (5) and newly added AnyAppBench templates (5), with three runs per template. Each cell averages 15 rollouts.}
\label{tab:app_level_music}
\AppLevelNoteRuns
\end{threeparttable}
\end{table*}

\begin{table*}[t]
\centering
\begin{threeparttable}
\AppTableSetup

\begin{tabularx}{\textwidth}{@{}ll*{5}{YZ}@{}}
\toprule
& & \multicolumn{2}{c}{\cellcolor{awshade}\makecell{\textbf{Simple Calendar}\\\textbf{Pro}}} & \multicolumn{2}{c}{\textbf{Etar}} & \multicolumn{2}{c}{\textbf{Fossify Calendar}} & \multicolumn{2}{c}{\textbf{Calendar}} & \multicolumn{2}{c}{\textbf{KashCal}} \\
\cmidrule(lr){3-4}\cmidrule(lr){5-6}\cmidrule(lr){7-8}\cmidrule(lr){9-10}\cmidrule(lr){11-12}
\textbf{Family} & \textbf{Model} & \textbf{AW} & \textbf{New} & \textbf{AW} & \textbf{New} & \textbf{AW} & \textbf{New} & \textbf{AW} & \textbf{New} & \textbf{AW} & \textbf{New} \\
\midrule
\multirow{5}{*}{\AppFamilyLabel{GUI\\Trained}} & UI-Venus-Navi-72B & 14.3 & 0.0 & 0.0 & 0.0 & 14.3 & 11.1 & 0.0 & 0.0 & 9.5 & 11.1 \\
 & UI-Venus-7B & 61.9 & 44.4 & 9.5 & 11.1 & 42.9 & 11.1 & 0.0 & 0.0 & 23.8 & 0.0 \\
 & GUI-Owl-7B & 4.8 & 0.0 & 0.0 & 0.0 & 14.3 & 0.0 & 0.0 & 0.0 & 0.0 & 0.0 \\
 & MAI-UI-8B & 4.8 & 0.0 & 9.5 & 0.0 & 71.4 & 77.8 & 23.8 & 0.0 & 38.1 & 11.1 \\
 & UI Voyager-4B & 33.3 & 44.4 & 9.5 & 0.0 & 28.6 & 11.1 & 28.6 & 0.0 & 23.8 & 11.1 \\
\midrule
\multirow{3}{*}{\AppFamilyLabel{General}} & Qwen3-VL-30B-A3B & 0.0 & 0.0 & 0.0 & 0.0 & 0.0 & 0.0 & 0.0 & 0.0 & 0.0 & 0.0 \\
 & Qwen3-VL-8B & 0.0 & 11.1 & 0.0 & 0.0 & 0.0 & 0.0 & 0.0 & 0.0 & 0.0 & 0.0 \\
 & Qwen2.5-VL-72B & 19.0 & 11.1 & 23.8 & 0.0 & 42.9 & 0.0 & 0.0 & 0.0 & 23.8 & 11.1 \\
\midrule
\multirow{2}{*}{\AppFamilyLabel{Closed}} & Gemini-3.1-Pro & 71.4 & 33.3 & 52.4 & 0.0 & 76.2 & 11.1 & 42.9 & 0.0 & 57.1 & 11.1 \\
 & GPT-5.1 & 61.9 & 11.1 & 9.5 & 11.1 & 76.2 & 33.3 & 42.9 & 0.0 & 42.9 & 11.1 \\
\midrule
\multirow{2}{*}{\AppFamilyLabel{Hybrid}} & Gemini-3.1-Pro$^\dagger$ & 66.7 & 33.3 & 52.4 & 0.0 & 57.1 & 11.1 & 42.9 & 0.0 & 42.9 & 0.0 \\
 & GPT-5.1$^\dagger$ & 66.7 & 33.3 & 28.6 & 0.0 & 71.4 & 44.4 & 14.3 & 0.0 & 14.3 & 0.0 \\
\midrule
\AppFamilyLabel{Agent} & Mobile-Agent-v3 & 66.7 & 66.7 & 0.0 & 0.0 & 81.0 & 33.3 & 0.0 & 0.0 & 28.6 & 0.0 \\
\midrule
\rowcolor{appavgrow}
\multicolumn{2}{l}{\textbf{Mean success rate}} & 36.3 & 22.2 & 15.0 & 1.7 & 44.3 & 18.8 & 15.0 & 0.0 & 23.4 & 5.1 \\
\bottomrule
\end{tabularx}
\AppLevelCaption{\textbf{App-level success rates for Calendar (\%, three runs pooled).} Each app has two sub-columns: AW-inherited templates (7) and newly added AnyAppBench templates (3), with three runs per template. Each cell averages 21 or 9 rollouts.}
\label{tab:app_level_calendar}
\AppLevelNoteRuns
\end{threeparttable}
\end{table*}

\begin{table*}[t]
\centering
\begin{threeparttable}
\AppTableSetup

\begin{tabularx}{\textwidth}{@{}ll*{4}{YZ}@{}}
\toprule
& & \multicolumn{2}{c}{\cellcolor{awshade}\makecell{\textbf{Simple SMS}\\\textbf{Messenger}}} & \multicolumn{2}{c}{\makecell{\textbf{Fossify}\\\textbf{Messages}}} & \multicolumn{2}{c}{\textbf{QUIK SMS}} & \multicolumn{2}{c}{\textbf{Messages}} \\
\cmidrule(lr){3-4}\cmidrule(lr){5-6}\cmidrule(lr){7-8}\cmidrule(lr){9-10}
\textbf{Family} & \textbf{Model} & \textbf{AW} & \textbf{New} & \textbf{AW} & \textbf{New} & \textbf{AW} & \textbf{New} & \textbf{AW} & \textbf{New} \\
\midrule
\multirow{5}{*}{\AppFamilyLabel{GUI\\Trained}} & UI-Venus-Navi-72B & 72.2 & 75.0 & 77.8 & 75.0 & 66.7 & 58.3 & 66.7 & 66.7 \\
 & UI-Venus-7B & 27.8 & 25.0 & 38.9 & 8.3 & 22.2 & 25.0 & 44.4 & 16.7 \\
 & GUI-Owl-7B & 72.2 & 58.3 & 66.7 & 50.0 & 38.9 & 25.0 & 77.8 & 50.0 \\
 & MAI-UI-8B & 100.0 & 66.7 & 72.2 & 58.3 & 88.9 & 66.7 & 83.3 & 83.3 \\
 & UI Voyager-4B & 94.4 & 25.0 & 77.8 & 16.7 & 61.1 & 25.0 & 72.2 & 8.3 \\
\midrule
\multirow{3}{*}{\AppFamilyLabel{General}} & Qwen3-VL-30B-A3B & 83.3 & 50.0 & 72.2 & 50.0 & 66.7 & 41.7 & 66.7 & 58.3 \\
 & Qwen3-VL-8B & 66.7 & 33.3 & 11.1 & 0.0 & 38.9 & 25.0 & 83.3 & 41.7 \\
 & Qwen2.5-VL-72B & 72.2 & 75.0 & 72.2 & 66.7 & 55.6 & 33.3 & 66.7 & 90.9 \\
\midrule
\multirow{2}{*}{\AppFamilyLabel{Closed}} & Gemini-3.1-Pro & 94.4 & 91.7 & 72.2 & 100.0 & 83.3 & 91.7 & 77.8 & 100.0 \\
 & GPT-5.1 & 94.4 & 91.7 & 83.3 & 83.3 & 94.4 & 58.3 & 88.9 & 91.7 \\
\midrule
\multirow{2}{*}{\AppFamilyLabel{Hybrid}} & Gemini-3.1-Pro$^\dagger$ & 88.9 & 100.0 & 88.9 & 100.0 & 94.4 & 100.0 & 94.4 & 100.0 \\
 & GPT-5.1$^\dagger$ & 83.3 & 91.7 & 100.0 & 91.7 & 83.3 & 75.0 & 88.9 & 90.9 \\
\midrule
\AppFamilyLabel{Agent} & Mobile-Agent-v3 & 100.0 & 33.3 & 100.0 & 25.0 & 44.4 & 33.3 & 77.8 & 33.3 \\
\midrule
\rowcolor{appavgrow}
\multicolumn{2}{l}{\textbf{Mean success rate}} & 80.8 & 62.8 & 71.8 & 55.8 & 64.5 & 50.6 & 76.1 & 64.0 \\
\bottomrule
\end{tabularx}
\AppLevelCaption{\textbf{App-level success rates for SMS (\%, three runs pooled).} Each app has two sub-columns: AW-inherited templates (6) and newly added AnyAppBench templates (4), with three runs per template. Each cell averages 18 or 12 rollouts.}
\label{tab:app_level_sms}
\AppLevelNoteRuns
\end{threeparttable}
\end{table*}

\begin{table*}[t]
\centering
\begin{threeparttable}
\AppTableSetup

\begin{tabularx}{\textwidth}{@{}ll*{5}{YZ}@{}}
\toprule
& & \multicolumn{2}{c}{\cellcolor{awshade}\makecell{\textbf{Material}\\\textbf{Files}}} & \multicolumn{2}{c}{\makecell{\textbf{Amaze File}\\\textbf{Manager}}} & \multicolumn{2}{c}{\makecell{\textbf{Fossify File}\\\textbf{Manager}}} & \multicolumn{2}{c}{\makecell{\textbf{Total}\\\textbf{Commander}}} & \multicolumn{2}{c}{\makecell{\textbf{X-plore File}\\\textbf{Manager}}} \\
\cmidrule(lr){3-4}\cmidrule(lr){5-6}\cmidrule(lr){7-8}\cmidrule(lr){9-10}\cmidrule(lr){11-12}
\textbf{Family} & \textbf{Model} & \textbf{AW} & \textbf{New} & \textbf{AW} & \textbf{New} & \textbf{AW} & \textbf{New} & \textbf{AW} & \textbf{New} & \textbf{AW} & \textbf{New} \\
\midrule
\multirow{5}{*}{\AppFamilyLabel{GUI\\Trained}} & UI-Venus-Navi-72B & 50.0 & 54.2 & 66.7 & 54.2 & 50.0 & 58.3 & 50.0 & 25.0 & 33.3 & 16.7 \\
 & UI-Venus-7B & 50.0 & 16.7 & 33.3 & 54.2 & 50.0 & 16.7 & 0.0 & 0.0 & 0.0 & 12.5 \\
 & GUI-Owl-7B & 50.0 & 37.5 & 16.7 & 25.0 & 0.0 & 25.0 & 16.7 & 0.0 & 16.7 & 16.7 \\
 & MAI-UI-8B & 83.3 & 58.3 & 100.0 & 66.7 & 100.0 & 70.8 & 83.3 & 41.7 & 33.3 & 54.2 \\
 & UI Voyager-4B & 50.0 & 29.2 & 50.0 & 33.3 & 50.0 & 25.0 & 50.0 & 0.0 & 16.7 & 20.8 \\
\midrule
\multirow{3}{*}{\AppFamilyLabel{General}} & Qwen3-VL-30B-A3B & 0.0 & 33.3 & 50.0 & 25.0 & 33.3 & 29.2 & 0.0 & 0.0 & 0.0 & 12.5 \\
 & Qwen3-VL-8B & 50.0 & 0.0 & 50.0 & 45.8 & 16.7 & 33.3 & 0.0 & 0.0 & 0.0 & 20.8 \\
 & Qwen2.5-VL-72B & 50.0 & 37.5 & 100.0 & 37.5 & 50.0 & 45.8 & 0.0 & 16.7 & 50.0 & 25.0 \\
\midrule
\multirow{2}{*}{\AppFamilyLabel{Closed}} & Gemini-3.1-Pro & 100.0 & 70.8 & 50.0 & 91.7 & 66.7 & 95.8 & 66.7 & 95.8 & 83.3 & 75.0 \\
 & GPT-5.1 & 50.0 & 37.5 & 66.7 & 37.5 & 83.3 & 87.5 & 66.7 & 20.8 & 0.0 & 12.5 \\
\midrule
\multirow{2}{*}{\AppFamilyLabel{Hybrid}} & Gemini-3.1-Pro$^\dagger$ & 100.0 & 100.0 & 100.0 & 87.5 & 100.0 & 100.0 & 100.0 & 100.0 & 100.0 & 100.0 \\
 & GPT-5.1$^\dagger$ & 100.0 & 83.3 & 83.3 & 58.3 & 100.0 & 95.8 & 50.0 & 75.0 & 50.0 & 58.3 \\
\midrule
\AppFamilyLabel{Agent} & Mobile-Agent-v3 & 50.0 & 20.8 & 83.3 & 50.0 & 50.0 & 54.2 & 50.0 & 16.7 & 50.0 & 33.3 \\
\midrule
\rowcolor{appavgrow}
\multicolumn{2}{l}{\textbf{Mean success rate}} & 60.3 & 44.6 & 65.4 & 51.3 & 57.7 & 56.7 & 41.0 & 30.1 & 33.3 & 35.3 \\
\bottomrule
\end{tabularx}
\AppLevelCaption{\textbf{App-level success rates for File Manager (\%, three runs pooled).} Each app has two sub-columns: AW-inherited templates (2) and newly added AnyAppBench templates (8), with three runs per template. Each cell averages 6 or 24 rollouts.}
\label{tab:app_level_file_manager}
\AppLevelNoteRuns
\end{threeparttable}
\end{table*}

\begin{table*}[t]
\centering
\begin{threeparttable}
\AppTableSetup

\begin{tabularx}{\textwidth}{@{}ll*{3}{YZ}@{}}
\toprule
& & \multicolumn{2}{c}{\cellcolor{awshade}\textbf{OsmAnd}} & \multicolumn{2}{c}{\textbf{Organic Maps}} & \multicolumn{2}{c}{\textbf{CoMaps}} \\
\cmidrule(lr){3-4}\cmidrule(lr){5-6}\cmidrule(lr){7-8}
\textbf{Family} & \textbf{Model} & \textbf{AW} & \textbf{New} & \textbf{AW} & \textbf{New} & \textbf{AW} & \textbf{New} \\
\midrule
\multirow{5}{*}{\AppFamilyLabel{GUI\\Trained}} & UI-Venus-Navi-72B & 0.0 & 0.0 & 22.2 & 33.3 & 0.0 & 28.6 \\
 & UI-Venus-7B & 0.0 & 4.8 & 44.4 & 33.3 & 11.1 & 42.9 \\
 & GUI-Owl-7B & 44.4 & 23.8 & 22.2 & 33.3 & 11.1 & 23.8 \\
 & MAI-UI-8B & 44.4 & 23.8 & 22.2 & 42.9 & 0.0 & 28.6 \\
 & UI Voyager-4B & 44.4 & 38.1 & 44.4 & 42.9 & 44.4 & 28.6 \\
\midrule
\multirow{3}{*}{\AppFamilyLabel{General}} & Qwen3-VL-30B-A3B & 0.0 & 38.1 & 0.0 & 28.6 & 0.0 & 38.1 \\
 & Qwen3-VL-8B & 33.3 & 42.9 & 22.2 & 57.1 & 0.0 & 52.4 \\
 & Qwen2.5-VL-72B & 11.1 & 42.9 & 0.0 & 38.1 & 0.0 & 23.8 \\
\midrule
\multirow{2}{*}{\AppFamilyLabel{Closed}} & Gemini-3.1-Pro & 66.7 & 71.4 & 22.2 & 81.0 & 44.4 & 61.9 \\
 & GPT-5.1 & 22.2 & 38.1 & 0.0 & 38.1 & 11.1 & 52.4 \\
\midrule
\multirow{2}{*}{\AppFamilyLabel{Hybrid}} & Gemini-3.1-Pro$^\dagger$ & 55.6 & 71.4 & 55.6 & 38.1 & 55.6 & 47.6 \\
 & GPT-5.1$^\dagger$ & 11.1 & 23.8 & 0.0 & 47.6 & 11.1 & 47.6 \\
\midrule
\AppFamilyLabel{Agent} & Mobile-Agent-v3 & 44.4 & 38.1 & 66.7 & 47.6 & 33.3 & 57.1 \\
\midrule
\rowcolor{appavgrow}
\multicolumn{2}{l}{\textbf{Mean success rate}} & 29.1 & 35.2 & 24.8 & 43.2 & 17.1 & 41.0 \\
\bottomrule
\end{tabularx}
\AppLevelCaption{\textbf{App-level success rates for Maps (\%, three runs pooled).} Each app has two sub-columns: AW-inherited templates (3) and newly added AnyAppBench templates (7), with three runs per template. Each cell averages 9 or 21 rollouts.}
\label{tab:app_level_maps}
\AppLevelNoteRuns
\end{threeparttable}
\end{table*}

\begin{table*}[t]
\centering
\begin{threeparttable}
\AppTableSetup

\begin{tabularx}{\textwidth}{@{}ll*{5}{YZ}@{}}
\toprule
& & \multicolumn{2}{c}{\cellcolor{awshade}\makecell{\textbf{Google}\\\textbf{Contacts}}} & \multicolumn{2}{c}{\makecell{\textbf{Fossify}\\\textbf{Contacts}}} & \multicolumn{2}{c}{\makecell{\textbf{Connect}\\\textbf{You}}} & \multicolumn{2}{c}{\makecell{\textbf{Simple Contacts}\\\textbf{Pro SE}}} & \multicolumn{2}{c}{\makecell{\textbf{Right}\\\textbf{Contact}}} \\
\cmidrule(lr){3-4}\cmidrule(lr){5-6}\cmidrule(lr){7-8}\cmidrule(lr){9-10}\cmidrule(lr){11-12}
\textbf{Family} & \textbf{Model} & \textbf{AW} & \textbf{New} & \textbf{AW} & \textbf{New} & \textbf{AW} & \textbf{New} & \textbf{AW} & \textbf{New} & \textbf{AW} & \textbf{New} \\
\midrule
\multirow{5}{*}{\AppFamilyLabel{GUI\\Trained}} & UI-Venus-Navi-72B & 100.0 & 66.7 & 50.0 & 4.2 & 0.0 & 0.0 & 100.0 & 70.8 & 0.0 & 8.3 \\
 & UI-Venus-7B & 50.0 & 12.5 & 50.0 & 0.0 & 0.0 & 0.0 & 100.0 & 25.0 & 0.0 & 0.0 \\
 & GUI-Owl-7B & 100.0 & 41.7 & 0.0 & 0.0 & 0.0 & 0.0 & 100.0 & 54.2 & 50.0 & 0.0 \\
 & MAI-UI-8B & 100.0 & 79.2 & 50.0 & 29.2 & 0.0 & 0.0 & 16.7 & 33.3 & 16.7 & 0.0 \\
 & UI Voyager-4B & 100.0 & 62.5 & 0.0 & 0.0 & 0.0 & 4.2 & 0.0 & 0.0 & 0.0 & 0.0 \\
\midrule
\multirow{3}{*}{\AppFamilyLabel{General}} & Qwen3-VL-30B-A3B & 83.3 & 75.0 & 50.0 & 8.3 & 0.0 & 8.3 & 50.0 & 37.5 & 0.0 & 0.0 \\
 & Qwen3-VL-8B & 0.0 & 29.2 & 50.0 & 4.2 & 0.0 & 0.0 & 50.0 & 16.7 & 0.0 & 8.3 \\
 & Qwen2.5-VL-72B & 100.0 & 50.0 & 0.0 & 0.0 & 0.0 & 4.2 & 50.0 & 16.7 & 0.0 & 0.0 \\
\midrule
\multirow{2}{*}{\AppFamilyLabel{Closed}} & Gemini-3.1-Pro & 100.0 & 100.0 & 100.0 & 91.7 & 50.0 & 70.8 & 100.0 & 87.5 & 100.0 & 95.8 \\
 & GPT-5.1 & 100.0 & 87.5 & 83.3 & 79.2 & 16.7 & 37.5 & 66.7 & 79.2 & 83.3 & 66.7 \\
\midrule
\multirow{2}{*}{\AppFamilyLabel{Hybrid}} & Gemini-3.1-Pro$^\dagger$ & 100.0 & 100.0 & 100.0 & 100.0 & 33.3 & 41.7 & 100.0 & 100.0 & 100.0 & 100.0 \\
 & GPT-5.1$^\dagger$ & 100.0 & 95.8 & 66.7 & 54.2 & 16.7 & 29.2 & 83.3 & 66.7 & 83.3 & 41.7 \\
\midrule
\AppFamilyLabel{Agent} & Mobile-Agent-v3 & 66.7 & 58.3 & 50.0 & 0.0 & 0.0 & 4.2 & 0.0 & 0.0 & 0.0 & 8.3 \\
\midrule
\rowcolor{appavgrow}
\multicolumn{2}{l}{\textbf{Mean success rate}} & 84.6 & 66.0 & 50.0 & 28.5 & 9.0 & 15.4 & 62.8 & 45.2 & 33.3 & 25.3 \\
\bottomrule
\end{tabularx}
\AppLevelCaption{\textbf{App-level success rates for Contacts (\%, three runs pooled).} Each app has two sub-columns: AW-inherited templates (2) and newly added AnyAppBench templates (8), with three runs per template. Each cell averages 6 or 24 rollouts.}
\label{tab:app_level_contacts}
\AppLevelNoteRuns
\end{threeparttable}
\end{table*}

\begin{table*}[t]
\centering
\begin{threeparttable}
\AppTableSetup

\begin{tabularx}{\textwidth}{@{}ll*{6}{YZ}@{}}
\toprule
& & \multicolumn{2}{c}{\cellcolor{awshade}\makecell{\textbf{Google}\\\textbf{Clock}}} & \multicolumn{2}{c}{\textbf{Clock}} & \multicolumn{2}{c}{\makecell{\textbf{Simple}\\\textbf{Clock}}} & \multicolumn{2}{c}{\makecell{\textbf{Clock}\\\textbf{You}}} & \multicolumn{2}{c}{\textbf{Chrono}} & \multicolumn{2}{c}{\makecell{\textbf{Fossify}\\\textbf{Clock}}} \\
\cmidrule(lr){3-4}\cmidrule(lr){5-6}\cmidrule(lr){7-8}\cmidrule(lr){9-10}\cmidrule(lr){11-12}\cmidrule(lr){13-14}
\textbf{Family} & \textbf{Model} & \textbf{AW} & \textbf{New} & \textbf{AW} & \textbf{New} & \textbf{AW} & \textbf{New} & \textbf{AW} & \textbf{New} & \textbf{AW} & \textbf{New} & \textbf{AW} & \textbf{New} \\
\midrule
\multirow{5}{*}{\AppFamilyLabel{GUI\\Trained}} & UI-Venus-Navi-72B & 66.7 & 81.0 & 0.0 & 14.3 & 66.7 & 71.4 & 66.7 & 14.3 & 33.3 & 57.1 & 66.7 & 71.4 \\
 & UI-Venus-7B & 66.7 & 90.5 & 0.0 & 0.0 & 66.7 & 47.6 & 0.0 & 0.0 & 66.7 & 57.1 & 77.8 & 47.6 \\
 & GUI-Owl-7B & 66.7 & 76.2 & 22.2 & 4.8 & 66.7 & 52.4 & 66.7 & 23.8 & 66.7 & 61.9 & 66.7 & 52.4 \\
 & MAI-UI-8B & 66.7 & 95.2 & 66.7 & 76.2 & 44.4 & 85.7 & 66.7 & 19.0 & 66.7 & 95.2 & 55.6 & 66.7 \\
 & UI Voyager-4B & 88.9 & 81.0 & 11.1 & 19.0 & 66.7 & 28.6 & 66.7 & 19.0 & 66.7 & 47.6 & 66.7 & 28.6 \\
\midrule
\multirow{3}{*}{\AppFamilyLabel{General}} & Qwen3-VL-30B-A3B & 55.6 & 90.5 & 44.4 & 57.1 & 66.7 & 57.1 & 66.7 & 14.3 & 66.7 & 47.6 & 66.7 & 61.9 \\
 & Qwen3-VL-8B & 77.8 & 76.2 & 77.8 & 81.0 & 66.7 & 61.9 & 66.7 & 14.3 & 66.7 & 57.1 & 66.7 & 38.1 \\
 & Qwen2.5-VL-72B & 66.7 & 81.0 & 22.2 & 19.0 & 66.7 & 81.0 & 66.7 & 28.6 & 44.4 & 71.4 & 66.7 & 81.0 \\
\midrule
\multirow{2}{*}{\AppFamilyLabel{Closed}} & Gemini-3.1-Pro & 100.0 & 100.0 & 44.4 & 0.0 & 66.7 & 85.7 & 66.7 & 23.8 & 0.0 & 0.0 & 66.7 & 85.7 \\
 & GPT-5.1 & 77.8 & 85.7 & 66.7 & 0.0 & 66.7 & 85.7 & 66.7 & 14.3 & 0.0 & 0.0 & 66.7 & 90.5 \\
\midrule
\multirow{2}{*}{\AppFamilyLabel{Hybrid}} & Gemini-3.1-Pro$^\dagger$ & 100.0 & 95.2 & 100.0 & 100.0 & 88.9 & 85.7 & 77.8 & 23.8 & 100.0 & 90.5 & 77.8 & 85.7 \\
 & GPT-5.1$^\dagger$ & 66.7 & 95.2 & 77.8 & 85.7 & 66.7 & 76.2 & 66.7 & 19.0 & 55.6 & 85.7 & 66.7 & 85.7 \\
\midrule
\AppFamilyLabel{Agent} & Mobile-Agent-v3 & 66.7 & 81.0 & 77.8 & 66.7 & 66.7 & 66.7 & 66.7 & 14.3 & 66.7 & 66.7 & 66.7 & 61.9 \\
\midrule
\rowcolor{appavgrow}
\multicolumn{2}{l}{\textbf{Mean success rate}} & 74.4 & 86.8 & 47.0 & 40.3 & 66.7 & 68.1 & 62.4 & 17.6 & 53.8 & 56.8 & 67.5 & 65.9 \\
\bottomrule
\end{tabularx}
\AppLevelCaption{\textbf{App-level success rates for Clock (\%, three runs pooled).} Each app has two sub-columns: AW-inherited templates (3) and newly added AnyAppBench templates (7), with three runs per template. Each cell averages 9 or 21 rollouts.}
\label{tab:app_level_clock}
\AppLevelNoteRuns
\end{threeparttable}
\end{table*}

\begin{table*}[t]
\centering
\caption{\textbf{AnyAppBench task and application inventory.}
Each of the 10 categories contains 10 task templates.
\textsuperscript{\tiny AW} marks AndroidWorld-derived task intents, bold app
names mark AndroidWorld-original canonical applications, and the remaining apps
are alternatives. The inventory contains 48 AndroidWorld-derived and 52 newly
added templates across 11 canonical and 41 alternative applications, yielding
520 category-matched task--application pairs.}
\label{tab:benchmark_overview_updated}

\small
\setlength{\tabcolsep}{3.2pt}
\renewcommand{\arraystretch}{1.05}

\begin{threeparttable}
\begin{tabularx}{\textwidth}{
@{}
>{\raggedright\arraybackslash}p{0.105\textwidth}
>{\raggedright\arraybackslash}X
>{\raggedright\arraybackslash}p{0.135\textwidth}
>{\raggedright\arraybackslash}p{0.215\textwidth}
>{\centering\arraybackslash}p{0.075\textwidth}
@{}
}
\rowcolor{tblblue}
\toprule
\textbf{Category}
& \textbf{Task Templates}
& \textbf{AW Orig. App(s)}
& \textbf{Newly Installed Apps}
& \makecell{\textbf{\# Apps}\\\textbf{AW+New}} \\
\midrule

Tasks / To-Do
& \awtask{CompletedTasksForDate}, \awtask{DueNextWeek}, \awtask{DueOnDate},
\awtask{HighPriorityTasks}, \awtask{HighPriorityDueOnDate}, \awtask{IncompleteTasksOnDate},
DueWithTime, Recurring, EditTask, CompleteTask
& \textbf{Tasks.org}
& Cfait, Todo List (PFA), ntodotxt, TaskMate, Grit
& \appsplit{1}{5} \\

\midrule

Notes
& \awtask{CreateNote}, \awtask{EditNote}, \awtask{MergeNotes},
\awtask{DeleteNote}, \awtask{SearchNote}, \awtask{ShareOrImportNote},
\awtask{FolderOrMoveNote}, \awtask{AttachOrTranscribeContent},
\awtask{CountTodoItems}, CreateChecklist
& \textbf{Joplin}, \textbf{Markor}
& NotallyX, neutriNote CE, Notesnook, Orgzly Revived
& \appsplit{2}{4} \\

\midrule

Finance
& \awtask{AddExpense}, \awtask{AddMultipleExpenses},
\awtask{DeleteTransaction}, \awtask{DeleteDuplicateTransactions},
\awtask{AttachReceipt}, AddIncome, EditTransaction,
CategorySummary, DateRangeTotal, TransferBetweenWallets
& \textbf{Pro Expense}
& Oinkoin, OpenMoneyBox, My Expenses, Finance Manager, Sushi
& \appsplit{1}{5} \\

\midrule

Music
& \awtask{CreatePlaylist}, \awtask{AddToPlaylist}, \awtask{AddToQueue},
\awtask{SaveOrExportPlaylist}, \awtask{PlaylistDuration},
RenamePlaylist, RemoveFromPlaylist, ReorderQueue,
SleepTimer, SearchAndPlay
& \textbf{Retro Music}
& Fossify Music Player, Apollo, SicMu Neo, Phonograph Plus, MonsterMusic
& \appsplit{1}{5} \\

\midrule

Calendar
& \awtask{AddOneEvent}, \awtask{AddTimedEvent}, \awtask{AddRepeatingEvent},
\awtask{DeleteEvent}, \awtask{EventsOnDate}, \awtask{NextEvent},
\awtask{EventsInRange}, EditEvent, AddReminder, MoveEvent
& \textbf{Simple Calendar Pro}
& Etar, Fossify Calendar, Calendar, KashCal
& \appsplit{1}{4} \\

\midrule

SMS
& \awtask{Send}, \awtask{Reply}, \awtask{ReplyMostRecent}, \awtask{Resend},
\awtask{SendClipboard}, \awtask{SendReceivedAddress},
CreateDraftMessage, EditDraftMessage,
DeleteConversation, ForwardMessage
& \textbf{Simple SMS Messenger}
& Fossify Messages, QUIK SMS, Messages
& \appsplit{1}{3} \\

\midrule

File Manager
& \awtask{DeleteFile}, \awtask{MoveFile},
CreateFolder, RenameFile, SaveCopyOfFile, SearchFile,
CompressFiles, ExtractArchive, ViewFileInfo, ShareFile
& \textbf{Material Files}
& Amaze File Manager, Fossify File Manager, Total Commander, X-plore File Manager
& \appsplit{1}{4} \\

\midrule

Maps
& \awtask{AddFavorite}, \awtask{AddMarker}, \awtask{RecordTrack},
SearchPlace, RemoveFavorite, DeleteMarker,
GetDirections, SearchNearbyPlace, ExportLocation, ShareLocation
& \textbf{OsmAnd\texttildelow}
& Organic Maps, CoMaps
& \appsplit{1}{2} \\

\midrule

Contacts
& \awtask{AddContact}, \awtask{NewContactDraft},
EditContact, SearchContact, ViewContactDetails,
AddFavoriteContact, RemoveFavoriteContact,
DeleteContact, CallContact, MessageContact
& \textbf{Google Contacts}
& Fossify Contacts, Connect You, Simple Contacts Pro SE, Right Contact
& \appsplit{1}{4} \\

\midrule

Clock
& \awtask{CreateTimer}, \awtask{StopwatchRunning}, \awtask{PauseStopwatch},
CreateAlarm, EditAlarm, EnableAlarm, DeleteAlarm,
StartTimer, StopwatchReset, AddWorldClock
& \textbf{Clock}
& Simple Clock, Google Clock, Clock You, Chrono, Fossify Clock
& \appsplit{1}{5} \\

\midrule
\rowcolor{tbltotal}
\textbf{Total}
& \textbf{48 AW-derived templates + \textcolor{green!45!black}{52 AnyAppBench-added templates}}
& \textbf{11 AW apps}
& \textbf{41 newly installed apps}
& \appsplit{11}{41} \\

\bottomrule
\end{tabularx}

\begin{tablenotes}[flushleft]
\footnotesize
\item Templates marked with \textsuperscript{\tiny AW} preserve AndroidWorld task intent, not necessarily the exact original AndroidWorld class name.
\item Unmarked templates are newly added AnyAppBench templates used to extend each category to 10 task templates.
\item \textbf{\# Apps AW+New} reports AndroidWorld-original apps plus newly installed apps; the number after ``+'' is the newly installed app count.
\end{tablenotes}
\end{threeparttable}
\end{table*}
\FloatBarrier
\subsection{Qualitative Examples}
\label{app:qualitative}

\begin{figure*}[t]
  \centering
  \includegraphics[width=\textwidth]{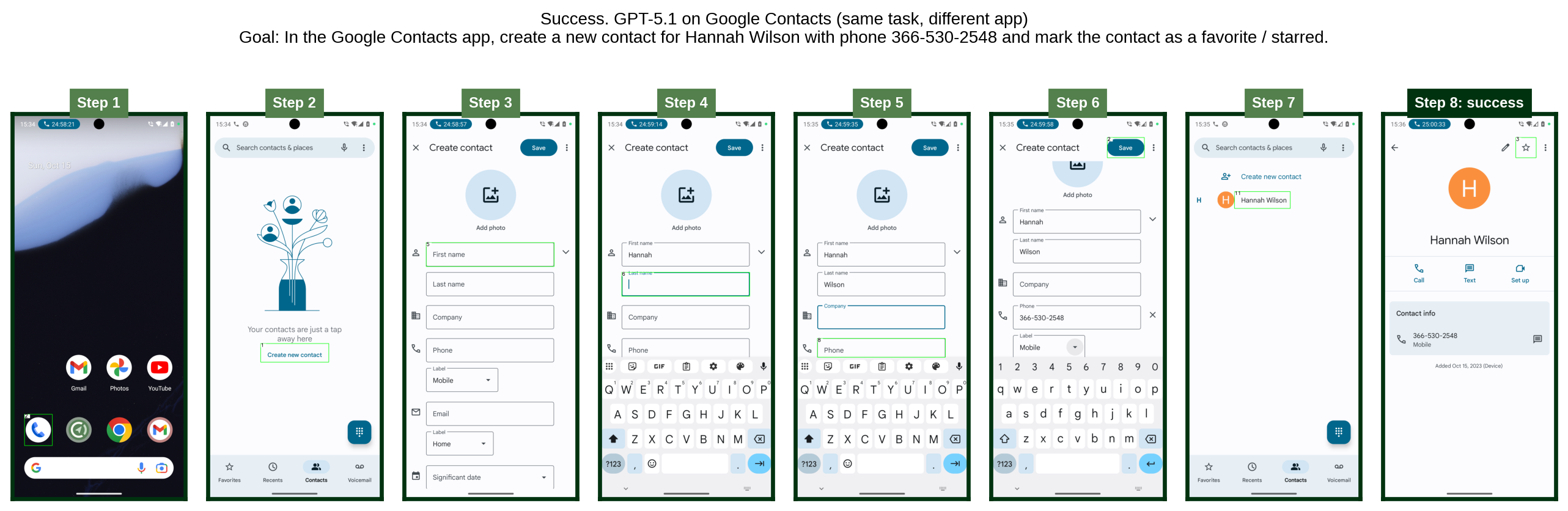}\\[4pt]
  \includegraphics[width=\textwidth]{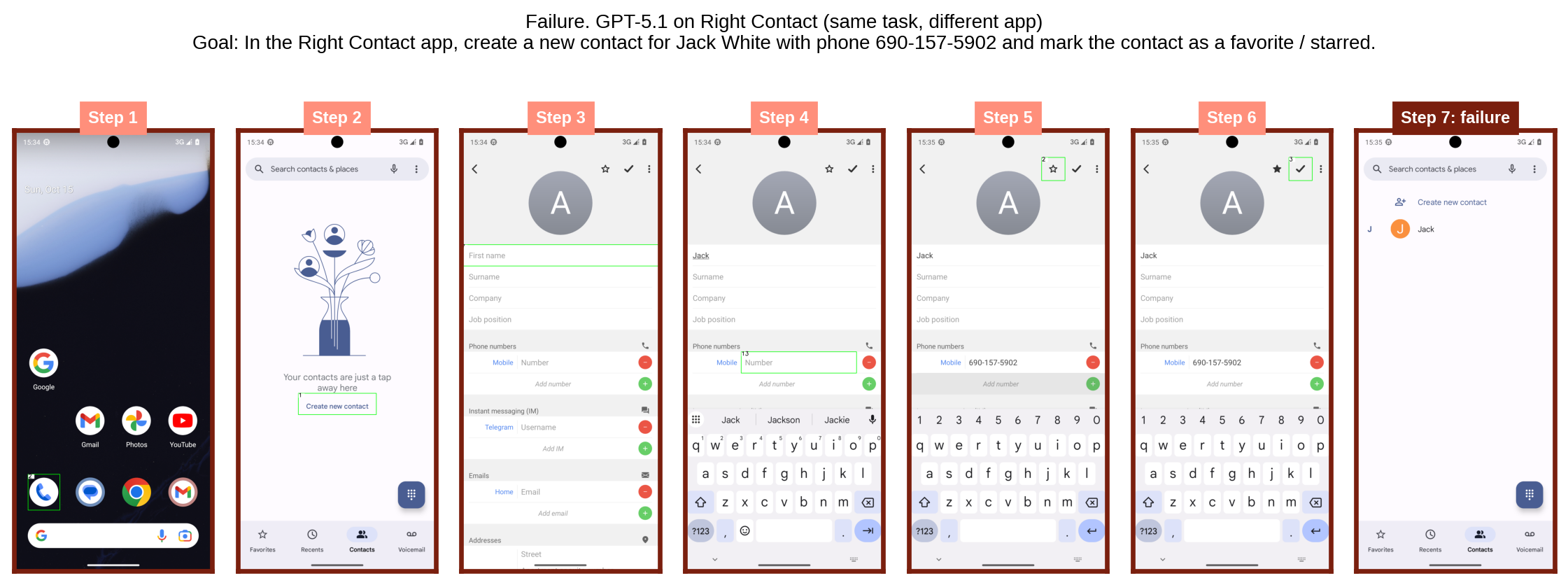}
  \caption{\textbf{Same task, different application.} GPT-5.1 performs the same
  semantic task (create a contact with a given phone number and mark it as a
  favorite) on two applications of the Contacts category. Top: on Google
  Contacts, the AndroidWorld-original app, the agent completes all fields,
  saves, and stars the contact (green blocks, verifier success). Bottom: on
  Right Contact, a new app, the agent enters the name and number but never
  completes the save and favorite flow (red blocks, verifier failure). The
  final frame of each strip carries the verifier outcome.}
  \label{fig:qualitative_examples}
\end{figure*}

Figure~\ref{fig:qualitative_examples} shows one successful and one failed episode of the same semantic task on two applications of the same category.

\clearpage
\raggedbottom
\FloatBarrier
\section{Additional RQ3 Results}
\label{app:rq3_full}

The main-text matched analysis covers all ten categories with two runs per condition. This appendix reports the per-application results of both category blocks.

The comparison on the remaining five categories (Tasks, Notes, Finance, Music, and Calendar) uses the same five agents and the same instruction-only Gemini-3.1-Pro generator. This cohort covers 29 applications and two matched runs per condition, giving 2{,}880 retained pairs after symmetric exclusion of ten incomplete cells. It was executed on a separate infrastructure, and every value was verified by an independent recomputation from the episode-level rollout ledger. On these categories the effect reverses. Success falls from 18.3\% under C1 to 14.8\% under C2, a change of $-$3.5 points (95\% CI $[-5.0, -2.0]$, McNemar $p<0.001$, 137 pairs gained and 238 lost). The drop is larger on the original AndroidWorld apps ($-$6.3 points) than on new apps ($-$2.8 points). The change is significant in Tasks, Notes, and Calendar and not in Finance or Music. Per agent, the change is $-$1.7 (UI-Venus-7B), $-$3.1 (GUI-Owl-7B), $-$0.9 (MAI-UI-8B), $-$1.0 (UI Voyager-4B), and $-$10.7 (Qwen3-VL-8B) points, and only the last is statistically significant.

\enlargethispage{20\baselineskip}
\noindent\begin{minipage}{\columnwidth}
\setlength{\parindent}{1em}
\indent Table~\ref{tab:rq3_per_app} reports the per-application C1 and C2 success rates and their difference for all ten categories, pooled over the matched agents and runs. Block (a) reconciles exactly with the aggregate values in the main text and block (b) with the aggregates above.
\end{minipage}

\begin{table*}[t]
\centering
\caption{\textbf{Per-application effect of sub-goal augmentation across all ten categories.} Success rate (\%) under C1 and C2 and the change, pooled over the five matched agents and the two matched runs of each cohort. $n$ counts matched rollout pairs. (AW) marks the AndroidWorld-original application of each category (Notes has two). Block (a) reconciles with Table~\ref{tab:rq3_c2_vs_c1_main} and block (b) with the aggregate statistics reported in this appendix.}
\label{tab:rq3_per_app}
\footnotesize
\setlength{\tabcolsep}{3.5pt}
\renewcommand{\arraystretch}{1.03}
\begin{adjustbox}{max width=\textwidth}
\begin{tabular}{lrrrr@{\hspace{18pt}}lrrrr}
\toprule
\multicolumn{5}{c}{\textbf{(a) SMS, File Manager, Maps, Contacts, Clock}} & \multicolumn{5}{c}{\textbf{(b) Tasks, Notes, Finance, Music, Calendar}} \\
\cmidrule(lr){1-5}\cmidrule(lr){6-10}
\textbf{Application} & \textbf{C1} & \textbf{C2} & $\boldsymbol{\Delta}$ & $\boldsymbol{n}$ & \textbf{Application} & \textbf{C1} & \textbf{C2} & $\boldsymbol{\Delta}$ & $\boldsymbol{n}$ \\
\midrule
\multicolumn{5}{l}{\emph{SMS}} & \multicolumn{5}{l}{\emph{Tasks/To-Do}} \\
Simple SMS Messenger (AW) & 59.0 & 68.0 & +9.0 & 100 & Tasks.org (AW) & 18.0 & 17.0 & -1.0 & 100 \\
Fossify Messages & 42.3 & 49.5 & +7.2 & 97 & Cfait & 9.0 & 4.0 & -5.0 & 100 \\
QUIK SMS & 45.0 & 55.0 & +10.0 & 100 & Grit & 6.0 & 7.0 & +1.0 & 100 \\
Messages & 58.0 & 67.0 & +9.0 & 100 & TaskMate & 11.0 & 5.0 & -6.0 & 100 \\
\quad All SMS & 51.1 & 59.9 & +8.8 & 397 & Todo List (PFA) & 10.0 & 11.0 & +1.0 & 100 \\
 & & & &  & ntodotxt & 6.0 & 1.0 & -5.0 & 100 \\
 & & & &  & \quad All Tasks/To-Do & 10.0 & 7.5 & -2.5 & 600 \\
\multicolumn{5}{l}{\emph{File Manager}} & \multicolumn{5}{l}{\emph{Notes}} \\
Material Files (AW) & 35.0 & 34.0 & -1.0 & 100 & Joplin (AW) & 34.0 & 20.0 & -14.0 & 100 \\
Amaze File Manager & 46.0 & 52.0 & +6.0 & 100 & Markor (AW) & 0.0 & 1.0 & +1.0 & 100 \\
Fossify File Manager & 33.3 & 53.5 & +20.2 & 99 & NotallyX & 35.0 & 31.0 & -4.0 & 100 \\
Total Commander & 13.0 & 12.0 & -1.0 & 100 & Notesnook & 36.0 & 32.0 & -4.0 & 100 \\
X-plore File Manager & 24.0 & 25.0 & +1.0 & 100 & Orgzly Revived & 33.0 & 40.0 & +7.0 & 100 \\
\quad All File Manager & 30.3 & 35.3 & +5.0 & 499 & neutriNote CE & 41.0 & 24.0 & -17.0 & 100 \\
 & & & &  & \quad All Notes & 29.8 & 24.7 & -5.2 & 600 \\
\multicolumn{5}{l}{\emph{Maps}} & \multicolumn{5}{l}{\emph{Finance}} \\
OsmAnd (AW) & 26.0 & 25.0 & -1.0 & 100 & Pro Expense (AW) & 34.0 & 29.0 & -5.0 & 100 \\
Organic Maps & 36.4 & 31.3 & -5.1 & 99 & Finance Manager & 27.0 & 29.0 & +2.0 & 100 \\
CoMaps & 27.6 & 37.8 & +10.2 & 98 & My Expenses & 18.8 & 15.6 & -3.1 & 96 \\
\quad All Maps & 30.0 & 31.3 & +1.3 & 297 & Oinkoin & 24.5 & 21.4 & -3.1 & 98 \\
 & & & &  & OpenMoneyBox & 5.0 & 7.0 & +2.0 & 100 \\
 & & & &  & Sushi & 20.0 & 13.0 & -7.0 & 100 \\
 & & & &  & \quad All Finance & 21.5 & 19.2 & -2.4 & 594 \\
\multicolumn{5}{l}{\emph{Contacts}} & \multicolumn{5}{l}{\emph{Music}} \\
Google Contacts (AW) & 48.0 & 70.0 & +22.0 & 100 & Retro Music (AW) & 10.0 & 5.0 & -5.0 & 100 \\
Fossify Contacts & 13.1 & 12.1 & -1.0 & 99 & Apollo & 0.0 & 1.0 & +1.0 & 98 \\
Connect You & 0.0 & 5.0 & +5.0 & 100 & Fossify Music & 5.1 & 9.2 & +4.1 & 98 \\
Simple Contacts Pro SE & 35.0 & 20.0 & -15.0 & 100 & MonsterMusic & 15.0 & 12.0 & -3.0 & 100 \\
Right Contact & 3.2 & 0.0 & -3.2 & 95 & Phonograph Plus & 0.0 & 0.0 & +0.0 & 100 \\
\quad All Contacts & 20.0 & 21.7 & +1.6 & 494 & SicMu Neo & 3.0 & 1.0 & -2.0 & 100 \\
 & & & &  & \quad All Music & 5.5 & 4.7 & -0.8 & 596 \\
\multicolumn{5}{l}{\emph{Clock}} & \multicolumn{5}{l}{\emph{Calendar}} \\
Google Clock (AW) & 79.0 & 81.0 & +2.0 & 100 & Simple Calendar Pro (AW) & 36.7 & 22.2 & -14.4 & 90 \\
Clock & 35.0 & 36.0 & +1.0 & 100 & Calendar & 15.0 & 13.0 & -2.0 & 100 \\
Simple Clock & 59.0 & 62.0 & +3.0 & 100 & Etar & 9.0 & 7.0 & -2.0 & 100 \\
Clock You & 27.0 & 24.0 & -3.0 & 100 & Fossify Calendar & 42.0 & 29.0 & -13.0 & 100 \\
Chrono & 65.0 & 59.0 & -6.0 & 100 & KashCal & 27.0 & 21.0 & -6.0 & 100 \\
Fossify Clock & 57.1 & 60.2 & +3.1 & 98 & \quad All Calendar & 25.7 & 18.4 & -7.3 & 490 \\
\quad All Clock & 53.7 & 53.7 & +0.0 & 598 &  & & & &  \\
\midrule
\textbf{All five categories} & \textbf{37.8} & \textbf{40.9} & \textbf{+3.2} & \textbf{2285} & \textbf{All five categories} & \textbf{18.3} & \textbf{14.8} & \textbf{-3.5} & \textbf{2880} \\
\bottomrule
\end{tabular}
\end{adjustbox}
\end{table*}

\vspace{0.5\baselineskip}
\FloatBarrier
\section{Disclosure of AI Assistance}
Following the ACL Policy on Publication Ethics, we disclose the generative-AI
tools used in preparing this work.

\paragraph{Writing assistance.} We used ChatGPT (OpenAI) only to improve the
grammar, fluency, and readability of author-drafted text. It was not used to
generate scientific content, research ideas, claims, analyses, or conclusions.

\paragraph{Coding assistance.} We used ChatGPT, OpenAI Codex, and Claude Code
(Anthropic) for boilerplate generation, refactoring, debugging, unit tests, and
visualization scripts. The authors designed the experiments, model architecture,
algorithms, and evaluation protocols, and reviewed, tested, and verified all
AI-assisted code before use.
\end{document}